\documentclass{article}
\pdfoutput=1 
\usepackage{arxiv}
\usepackage[linesnumbered,ruled,vlined]{algorithm2e}

\AtBeginDocument{%
  \providecommand\BibTeX{{%
    \normalfont B\kern-0.5em{\scshape i\kern-0.25em b}\kern-0.8em\TeX}}}

\usepackage{multirow}
\usepackage{hhline}
\usepackage{framed}
\usepackage{xspace}
\usepackage[]{mdframed}
\usepackage{listings}
\usepackage{tcolorbox}

\usepackage{color, colortbl}
\usepackage{xcolor}
\definecolor{LightCyan}{rgb}{0.85,0.85,0.85}
\definecolor{DarkCyan}{rgb}{0.9,0.9,0.9}
\definecolor{DarkGreen}{rgb}{0,0.5,0}
\definecolor{LightGreen}{rgb}{0.8,0.95,0.8}
\definecolor{LightRed}{rgb}{0.95,0.8,0.8}
\definecolor{LightBlue}{rgb}{0.68,0.85,0.90}
\usepackage{booktabs}
\usepackage{enumitem}
\usepackage{amsmath}
\usepackage{hyperref}
\usepackage{cleveref}
\usepackage{amsfonts}
\usepackage{xfrac}
\usepackage[toc]{appendix}
\usepackage{adjustbox}
\usepackage{subcaption}
\usepackage{graphicx}
\usepackage{amssymb}

\crefname{section}{§}{§§}
\begin{document}

\title{TESSERACT: Eliminating Experimental Bias in Malware Classification across Space and Time (Extended Version)}

\author{Zeliang Kan \\
  King's College London \\
  University College London \\
  HiddenLayer
  \And
  Shae McFadden \\
  King's College London \\
  The Alan Turing Institute \\
  University College London \\
  \And
  Daniel Arp \\
  TU Wien \\
  \And
  Feargus Pendlebury \\
  University College London \\
  \And
  Roberto Jordaney \\
  Royal Holloway,\\University of London \\
  \And
  Johannes Kinder \\
  LMU Munich \\
  \And 
  Fabio Pierazzi \\
  University College London \\
  King's College London \\
  \And 
  Lorenzo Cavallaro \\
  University College London
}

\renewcommand{\shorttitle}{}
\renewcommand{\headeright}{TESSERACT (Extended Version)}
\renewcommand{\undertitle}{}
\newcommand{\tesseract}{\textsc{Tesseract}\xspace}

\newcommand{\sklearn}{\textsf{\textsc{Scikit-learn}}}
\newcommand{\torch}{\textsf{\textsc{PyTorch}}}

\newcommand{\eg}{e.g.,}
\newcommand{\ie}{i.e.,}

\newcommand{\drebin}{\textsc{Drebin}\xspace}
\newcommand{\dl}{\textsc{DeepDrebin}\xspace}
\newcommand{\mmd}{\textsc{MaMaDroid}\xspace}
\newcommand{\transcend}{\textsc{Transcendent}\xspace}
\newcommand{\ember}{\textsc{EMBER}\xspace}
\newcommand{\hidost}{\textsc{Hidost}\xspace}

\newcommand{\etal}{\textit{et al.}}

\newcommand{\dataset}{$D$}
\newcommand{\labelingcost}{$L_c$}
\newcommand{\quarantinecost}{$Q_c$}
\newcommand{\performance}{P}
\newcommand{\performancestar}{$P^*$}

\newcommand{\errorrate}{$E$}
\newcommand{\errorratemax}{$E_{max}$}

\maketitle

\begin{abstract}

  Machine learning (ML) plays a pivotal role in detecting malicious software. Despite the high $F_1$-scores reported in numerous studies reaching upwards of 0.99, the issue is not completely solved. Malware detectors often experience performance decay due to constantly evolving operating systems and attack methods, which can render previously learned knowledge insufficient for accurate decision-making on new inputs. This paper argues that commonly reported results are inflated due to two pervasive sources of experimental bias in the detection task: \textit{spatial bias} caused by data distributions that are not representative of a real-world deployment; and \textit{temporal bias} caused by incorrect time splits of data, leading to unrealistic configurations. To address these biases, we introduce a set of constraints for fair experiment design, and propose a new metric, AUT, for classifier robustness in real-world settings. We additionally propose an algorithm designed to tune training data to enhance classifier performance. Finally, we present \tesseract, an open-source framework for realistic classifier comparison. Our evaluation encompasses both traditional ML and deep learning methods, examining published works on an extensive Android dataset with 259,230 samples over a five-year span. Additionally, we conduct case studies in the Windows PE and PDF domains. Our findings identify the existence of biases in previous studies and reveal that significant performance enhancements are possible through appropriate, periodic tuning. We explore how mitigation strategies may support in achieving a more stable and better performance over time by employing multiple strategies to delay performance decay.

\end{abstract}


\section{Introduction}
\label{sec:intro}

Machine learning (ML) has become a fundamental tool in malware research within the academic security community, with its applications spanning a wide range of domains. These include Windows malware \cite{Dahl:Large,Tahan:Mal,Markel:Building, Endgame:Ember}, PDF malware \cite{Laskov:PDF,Maiorca:PDF, vsrndic2016hidost}, malicious URLs \cite{Stringhini:Shady,Lee:Warningbird}, malicious JavaScript \cite{Rieck:Cujo,Curtsinger:Zozzle}, and Android malware \cite{Arp:Drebin,Mariconti:MaMaDroid,Papernot:ESORICS, zhang2020enhancing}. While the high performance figures associated with machine learning solutions might lead one to believe that the issue of malware classification is practically solved, this is far from true.

Malware classifiers are challenged by the dynamic nature of operational environments and attacking techniques. The constant emergence of new malware variants and families often leads to a degradation in the predictive accuracy of these classifiers over time \cite{Mariconti:MaMaDroid}. Ensuring temporal consistency is therefore a critical factor in evaluating the effectiveness of these classifiers. A skewed outcome can result if the experimental setup inadvertently provides the classifier with foreknowledge, leading to biased results \cite{Allix:Timeline, Miller:Reviewer}.

These biases are not isolated instances but are widespread in different security domains. In this paper, we primarily focus on the domain of Android malware, arguing that many detection approaches, including those cited \cite{Arp:Drebin,Suarez:DroidSieve,Mariconti:MaMaDroid,Gascon:Structural,Zhang:Semantics,Dash:Droidscribe,Yuan:DroidSec,Papernot:ESORICS} and even encompassing our previous research, are often evaluated under conditions that do not accurately reflect real-world scenarios. Then in order to broaden our perspective, we also extend our investigation into the Windows and PDF malware domains by case study \cite{Endgame:Ember,vsrndic2016hidost}.

We identify two types of experimental bias: \textit{spatial bias} and \textit{temporal bias}. \textit{Spatial bias} occurs when there are unrealistic assumptions about the ratio of goodware to malware in the dataset, which must be consistently enforced during the testing phase to mimic a realistic scenario. The ratio of goodware to malware varies by domain. For example, most Android apps in the wild are goodware~\cite{lindorfer2014andradar,Google:Report,Allix:AndroZoo}, whereas most URLs in software download events are malicious~\cite{Maggi:URLs, Perdisci:URLs}. \textit{Temporal bias} occurs when there is a temporal inconsistency between the training and testing datasets, which can lead to the integration of future knowledge into the training phase or create unrealistic settings, especially for concept drift detection~\cite{Allix:Timeline,Miller:Reviewer,barbero2022transcending}. This issue is particularly concerning in the case of families of closely related malware, where including even one variant in the training set may enable the algorithm to identify many variants in the testing.

We believe that the pervasiveness of these issues is due to two main reasons: first, possible sources of evaluation bias are not common knowledge; second, accounting for time complicates the evaluation and does not allow a comparison to other approaches using headline evaluation metrics such as the $F_1$-Score or AUROC. We address these issues in this paper by systematizing evaluation bias for Android malware classification and providing new constraints for sound experiment design along with new metrics and framework support.

Prior work has investigated challenges and experimental bias in security evaluations~\cite{rossow2012prudent,Sommer:Outside,Van:BenchmarkingCrimes, Allix:Timeline, Miller:Reviewer, Axelsson:BaseRate}. The \emph{base-rate fallacy}~\cite{Axelsson:BaseRate} describes how evaluation metrics such as True Positive Rate ($\mathit{TPR}$) and False Positive Rate ($\mathit{FPR}$) are misleading in intrusion detection, due to significant class imbalance (e.g., most traffic is benign); in contrast, we identify and address experimental settings that give misleading results \emph{regardless} of the adopted metrics---even when correct metrics are reported. Previous works by Sommer and Paxson~\cite{Sommer:Outside}, Rossow \etal~\cite{rossow2012prudent}, and Kouwe \etal~\cite{Van:BenchmarkingCrimes} propose guidelines for sound security evaluations, but none of them identify temporal and spatial bias or quantify the impact of errors on classifier performance. While Allix \etal~\cite{Allix:Timeline} and Miller \etal~\cite{Miller:Reviewer} identify initial temporal constraints in Android malware classification, we demonstrate that recent work following their guidelines (e.g.,~\cite{Arp:Drebin, Endgame:Ember}) still suffer from temporal and spatial bias. Our paper is the first to identify and address these sources of bias with novel, actionable constraints, metrics, and tool support.

This article makes the following contributions:
\begin{itemize}

\item We identify two sources of bias that affect the evaluation of ML-based malware classifiers: \emph{temporal} bias associated with incorrect train-test splits and \emph{spatial} bias related to unrealistic assumptions in dataset distribution. As an extension of our previous conference paper, we experimentally verify our hypothesis on an extended Android dataset of 259,230 apps (with 10\% malware) spanning over five years that, due to bias, performance can decrease up to 50\% in practice in the well-known Android malware classifiers. Furthermore, we explore the impact of these biases on popular classifiers in other security domains, specifically targeting Windows PE \cite{Endgame:Ember} and PDF malware \cite{vsrndic2016hidost}.

\item We propose novel building blocks for more robust evaluations of malware classifiers: a set of spatio-temporal constraints to be enforced in experimental settings; a new metric, AUT, that captures a classifier's robustness to time decay with optional evaluation parameters, and allows for fair comparison of different algorithms; and an updated tuning algorithm that empirically optimizes the classification performance when malware represents the minority class. In comparing performances across Android, Windows PE, and PDF malware domains, we demonstrate how eliminating biases can provide counter-intuitive results on real performance. Additionally, our findings underscore the necessity of regular retraining and re-tuning to maintain stable classifier performance. 

\item We implement and publicly release the Python framework for our methodology, \tesseract. Accompanied by a comprehensive user guide. It demonstrates how our findings evaluate the performance-cost trade-offs of solutions designed to mitigate time decay, such as active learning \cite{Settles:AL} and conformal prediction (CP) \cite{saunders1998ridge}. Additionally, we implement the state-of-the-art CP framework-\transcend \cite{barbero2022transcending}-to be compatible with deep neural networks (DNN). This adaptation involves the implementation of a non-conformity measurement (NCM) based on the output from the SoftMax layer of a DNN model.

\end{itemize}

This article extends our work published at USENIX Security Symposium 2019~\cite{pendlebury2019tesseract}. In this journal version, we focus on deepening and expanding the most promising aspects observed in our initial study on a larger dataset. A key enhancement is our exploration of the deep learning-based malware detection approach, previously referred to as `DL' in \cite{pendlebury2019tesseract} and detailed in Appendix~\ref{app:hyperparameters}. This version presents an in-depth analysis of the deep learning approach, particularly examining how it performs in comparison to the traditional ML method when applied to identical spatial and temporal bias. 
Furthermore, we extend our exploration to additional security domains, specifically Windows PE and PDF malware. Our findings reveal that, even in these domains where concept drift is less pronounced, applying our proposed tuning algorithm to the training data can still enhance detection capabilities.

Our \tesseract framework offers the research community a tool for producing consistent results, uncovering unexpected performance insights, and assessing classifiers in real-world industrial scenarios. While we make strides in understanding spatio-temporal biases in experimental setups, there remains a vast expanse of unexplored territory. The potential to further investigate diverse security domains, encompassing a range of malware families, operating systems, and digital environments, is substantial. We advocate for the broader security community to engage in this exploration, building upon the foundational principles we have laid out.

The structure of this article is as follows: In \S~\ref{sec:Android Malware Classification}, we present the background knowledge pertinent to our Android malware classification evaluation. Subsequently, in \S~\ref{sec:bias}, we identify and validate potential sources of experimental bias. Our methodology for conducting a spatio-temporal bias-free evaluation is delineated in \S~\ref{sec:Space-Time Aware Evaluation}. In \S~\ref{sec:otherdomains}, we assess the influence of these biases on the Windows PE and PDF domains by case study. \S~\ref{sec:delay} demonstrates how \tesseract can be utilized to compare and assess the trade-offs associated with various budget-constrained strategies to mitigate time decay. Then, we discuss related research in \S~\ref{sec:related}. Lastly, we discuss guidelines, assumptions, limitations, and availability in \S~\ref{sec:discussion} and \S~\S~\ref{sec:availability}, and make conclusions in \S~\ref{sec:conclusions}.

\begin{tcolorbox}
\textbf{Note: Use of the term ``bias''} In this article, the term \emph{(experimental) bias} is employed to denote the discrepancies between an experimental setting and the conditions observed in genuine real-world deployments. These discrepancies might give rise to evaluations that are potentially misleading. It is pivotal to differentiate that we are not referencing the trade-off between classifier bias and variance as delineated in the conventional machine learning literature~\cite{Bishop:ML}.
\end{tcolorbox}
\section{Android Malware Classification}
\label{sec:Android Malware Classification}
We first delve into the Android domain. In this section, we outline the reference of examined Android classification approaches in \S~\ref{subsec:Reference Algorithms}. We also discuss the specific prevalence of malware within the Android domain in \S~\ref{subsec:Malware Ratio}, detail the composition of the Android dataset under validation in \S~\ref{subsec:Dataset}, and elaborate on our chosen features and feature selection methodology \S~\ref{subsec:select}.

\subsection{Reference Algorithms in Android Classification}
\label{subsec:Reference Algorithms}

In this study, we primarily focus on two popular malware detection approaches: \drebin and \dl. We opt for \drebin's approach and its feature because it is a highly regarded technique in state-of-the-art machine learning security research \cite{barbero2022transcending, daoudi2022deep, chen2023continuous}. It utilizes a linear support vector machine (SVM) with high-dimensional binary feature vectors derived from efficient static analysis. Additionally, we incorporate \dl~\cite{Papernot:ESORICS}, a deep learning approach that uses the same input features as \drebin. Deep learning models are known for their ability to discover latent feature spaces, thus potentially offering enhanced resilience against time decay~\cite{Goodfellow:DL}. Our previous work, published at USENIX Security 2019 \cite{pendlebury2019tesseract}, indicated that deep learning frameworks could outperform traditional machine learning methods in terms of robustness, it becomes interesting to investigate whether these initial findings persist under varied evaluation scenarios.

To ensure a robust experimental framework, we have thoroughly re-implemented both \drebin and \dl based on the detailed methodologies outlined in their foundational papers \cite{Arp:Drebin,Papernot:ESORICS}. Successfully replicating baseline results from these studies highlights their scientific rigor and forms the foundation upon which we build our research. These methods are selected for their accessibility and the stable baselines they provide, allowing us to concentrate on the implications of our findings within the wider malware detection research field. The hyper-parameters of the re-implemented algorithms are detailed in Appendix~\ref{app:hyperparameters}. In contrast, we have excluded \mmd~\cite{Mariconti:MaMaDroid} from this study due to its limited efficacy, as observed in our previous analysis spanning 2014 to 2016. Our findings indicated that \mmd's performance had plateaued, making further exploration of its Markov-chain feature inefficient.

\subsection{Estimating in-the-wild Malware Ratio}
\label{subsec:Malware Ratio}

The proportion of malware in a dataset can have a significant impact on the performance of a classifier (see \S~\ref{sec:bias}). In order to conduct unbiased experiments, it is crucial to use a dataset with a realistic percentage of malware to goodware. This can be achieved by down-sampling the majority class, for instance. However, it is important to note that each malware domain has its own unique ratio of malware to goodware that is typically encountered in the wild. It is therefore essential to determine whether malware is a minority, majority, or equal-sized class compared to goodware. For example, malware is the minority class in network traffic~\cite{Axelsson:BaseRate} and Android~\cite{lindorfer2014andradar}, but it is the majority class in binary download events~\cite{Perdisci:URLs}. Estimating the percentage of malware in the wild for a given domain can be challenging, but it can be informed by measurement papers, industry telemetry, and publicly available reports.

In the Android ecosystem, there is variation in the proportion of malware among all the apps. For instance, industry telemetry by a key player suggests that around 6\% of Android apps are malicious. On the other hand, AndRadar measurement study~\cite{lindorfer2014andradar} indicates that the ratio of Android malware in the wild is approximately 8\%. The 2017 Google’s Android security report~\cite{Google:Report} estimates the malware percentage to be between 6\% and 10\%, while in 2023 the AndroZoo~\cite{Allix:AndroZoo} dataset analysis reveals an incidence of 16.3\% of malicious apps among the whole dataset.
This data suggests that, on average, malware is the minority class in the Android domain. To stabilize the percentage of malware in our dataset, we choose to set it to 10\%, which is an approximate average across different estimates, with monthly values ranging from 8\% to 12\%. This decision allows us to collect a dataset with statistically significant per-month malware samples. Aggressive under-sampling of the majority class would have decreased the statistical significance of the dataset, while oversampling of the goodware class would have been too resource-intensive. (see \S~\ref{subsec:Dataset})

It is important to emphasize that the decision to set the malware percentage at 10\% in our dataset should not be interpreted as a universal benchmark for Android malware detection. The landscape of malware and software development is dynamic, with constantly evolving threats and defensive measures. Therefore, researchers and practitioners should exercise discretion in selecting an appropriate malware-to-goodware ratio for their specific context. This selection should be informed by the latest trends, industry reports, and the specific characteristics of the domain being studied. Regular reassessment of this ratio is recommended to ensure the continued relevance and accuracy of the malware detection methodologies being employed.

\subsection{Android Dataset}
\label{subsec:Dataset}

In our research, we utilize AndroZoo~\cite{Allix:AndroZoo} as our data source, which is publicly available and contains more than 22.6 million Android apps when we write the article. Each sample from AndroZoo is labeled with a timestamp and many of them also include VirusTotal metadata results. This dataset is constantly updated through crawling from various markets, including Google Play Store and third-party markets such as Anzhi and AppChina. We selected AndroZoo dataset for our experiments because of its large size and extensive time-span, which allows us to perform realistic experiments that are aware of both space and time factors.

\textbf{Defining goodware and malware.} Ideally, goodware in software security refers to legitimate Android apps that do not contain any malicious behavior. Malware, on the other hand, refers to those that have been classified as containing malicious behavior. However, different anti-virus engines may have different criteria for what is malicious. Since AndroZoo’s metadata reports the number $p$ of positive anti-virus reports on VirusTotal~\cite{VT:URL} for samples in the AndroZoo dataset, we chose $p = 0$ for goodware and $p \geq 4$ for malware, following Miller \etal{}’s~\cite{Miller:Reviewer} advice for a reliable ground-truth. About 9\% of AndroZoo apps can be called grayware as they have $0 < p < 4$ when we write the article. We exclude grayware when constructing our dataset, as including gray samples could disadvantage classifiers whose features were designed with a different labeling threshold.

\textbf{Time-span selection.} In this extended article, one of our goals is to validate our previous findings by employing a larger, more comprehensive Android dataset. The vastness of the complete AndroZoo dataset, which total size of DEX files stands at over 127TB, coupled with the constraints of our research infrastructure, poses significant challenges in extracting features from the entire dataset. For context, utilizing our research infrastructure — which consists of three high-end Dell PowerEdge R730 nodes (boasting 2 x 14 cores in hyperthreading, leading to a total of 168 vCPU threads, complemented by 1.2TB of RAM and an accompanying 100TB NAS)—to extract \drebin's features for the entire dataset is estimated to take close to three years. 

Practical considerations, including the costs of feature extraction and storage space requirements, drive our decision to find a balance between the temporal coverage and representativeness of our dataset. We select the period from 2014 to 2018 for analysis, a timeframe frequently cited in the assessment of contemporary methodologies, as highlighted in \cite{chen2023overkill, barbero2022transcending, chen2023continuous}. This choice, we believe, provides substantial statistical significance, crucial for identifying and thoroughly evaluating the influence of experimental biases on model performance. In our examination of time decay within the Android domain, we adopt a monthly granularity and uniformly sample 259K AndroZoo apps from January 2014 to December 2018. Furthermore, we maintain an overall malware average of 10\%, in line with the malware ratio around 2017 as detailed in \S~\ref{subsec:Malware Ratio}, while allowing a monthly variation between 8\% and 12\% to ensure some degree of fluctuation. This five-year span ensures a robust dataset, offering over 1,000 apps per month, with the exception of the final quarter of 2016, which experienced a lower application crawl rate.

\begin{figure}[t]
 	\includegraphics[width=1.0\columnwidth]{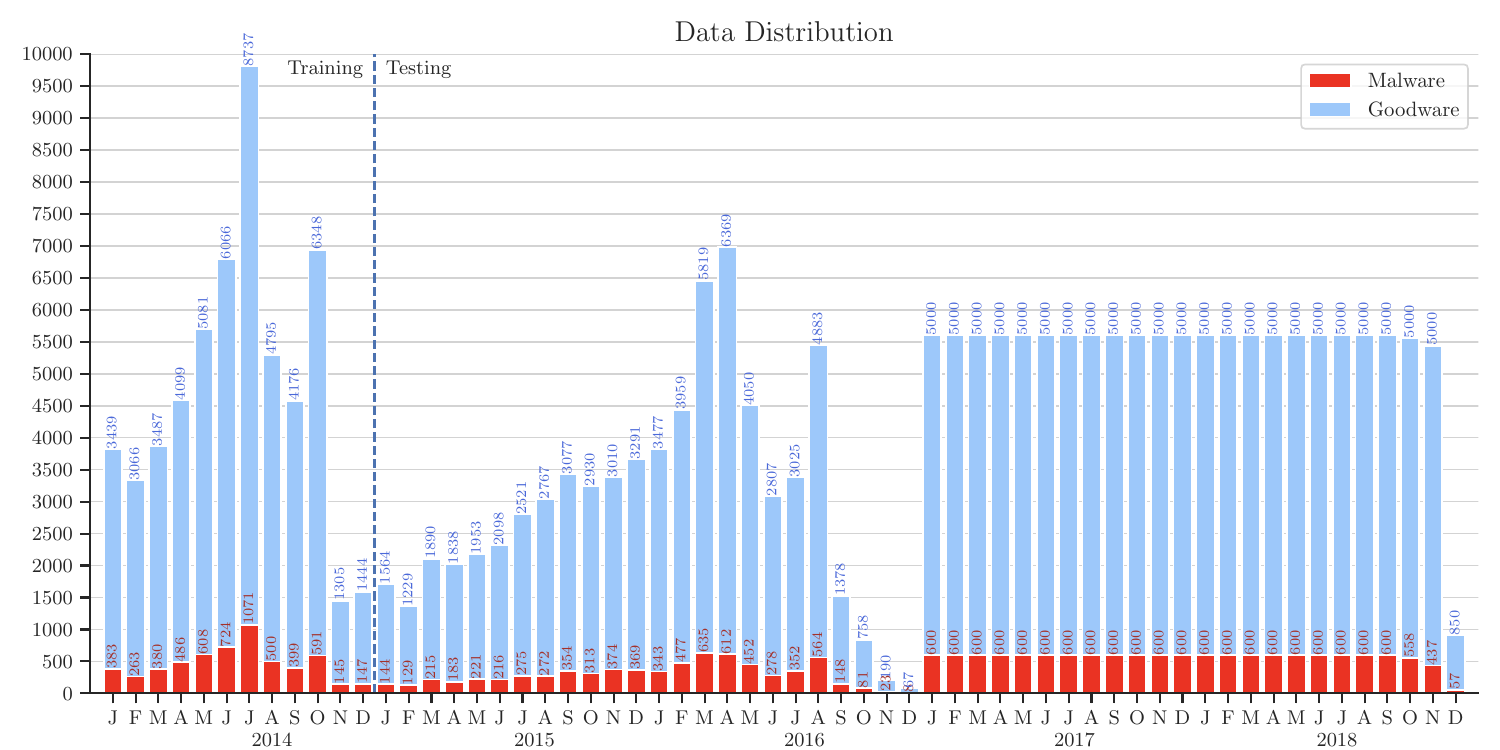}
 	\caption{Data distribution of the Android dataset using for this study. The figure shows a stacked histogram illustrating the monthly distribution of Android APKs we sourced from AndroZoo. It comprises $259,230$ Android applications, with approximately 10\% being malware each month, covering a period from Jan. 2014 to Dec. 2018. The vertical dashed line indicates the split for all time-aware experiments in this study, with training data from 2014 and testing data from 2015 to 2018 if there is no special illustration. }
 	\label{fig:dataset}
\end{figure}

\textbf{Dataset summary.} The final dataset used in our study consists of 259,230 Android applications, with 232,843 labeled as goodware and 26,387 labeled as malware. To visualize the distribution of goodware and malware over time, we provide a stack histogram in \autoref{fig:dataset}. The histogram displays the per-month distribution, and each bar represents a specific time period. For clarity, the figure also indicates the number of malware and goodware samples in each bar. If not explicitly stated, all subsequent time-aware experiments discussed in this paper involve training on data from 2014 and testing on data from 2015 to 2018. This time range is indicated by the vertical dashed line in \autoref{fig:dataset}, illustrating the division between the training and testing periods.

\subsection{Feature Selection}
\label{subsec:select}

We also confront the challenge of a large \drebin~\cite{Arp:Drebin} feature space present in the dataset mentioned earlier, which encompasses over two million features across eight different aspects of Android apps, including permissions, intents, API calls, among others \cite{Arp:Drebin}. This large feature space significantly increases computational demands and extends experiment duration, particularly problematic when analyzing \dl in detail. Owing to \dl's architecture, such a vast number of two million input neurons leads to an exponential increase in the number of parameters, resulting in prolonged training/testing times.

To address this, we employ feature selection techniques that aim to reduce the size of the feature space while maintaining its effectiveness in discriminating between different samples. Previous research studies~\cite{demontis2017yes, Roy:ExpML, chen2023continuous} have shown that utilizing only the top 10,000 features does not significantly affect \drebin classification performance. This is due to only a small number of features displaying significant discrimination and being assigned non-zero weights by the SVM learning algorithm used in \drebin classification.

We begin by initializing the \drebin approach on the full feature space with data from 2014. Since the chosen classification algorithm is a linear SVM, the decision function after training will be $f(x) = \mathbf{w}^{T}\mathbf{x} = \sum_{i=1}^{n} w_ix_i$. Here, $n$ is the dimension of original feature space, $\mathbf{w}$ is the model's weight vector, and $\mathbf{x}$ represents the input sample. According to the algorithm, the absolute value of each $w_i$ in its weight vector is positively correlated with the importance of the $i^{th}$ feature in making a classification decision.

\begin{figure}[t]
\centering
	\begin{subfigure}{0.23\textwidth}
		\centering
		\includegraphics[width=\linewidth]{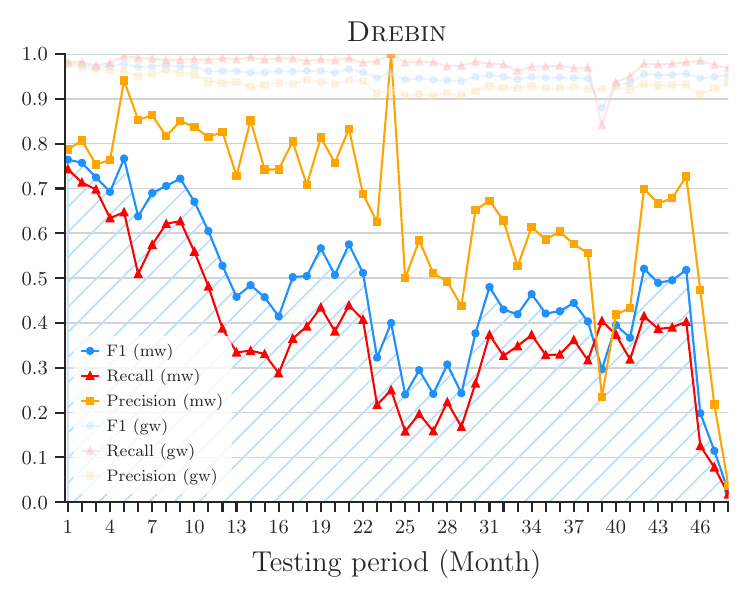}
		\caption{All features}
        \label{fig:reduce-drebin-full}
	\end{subfigure}
	\begin{subfigure}{0.23\textwidth}
		\centering
		\includegraphics[width=\linewidth]{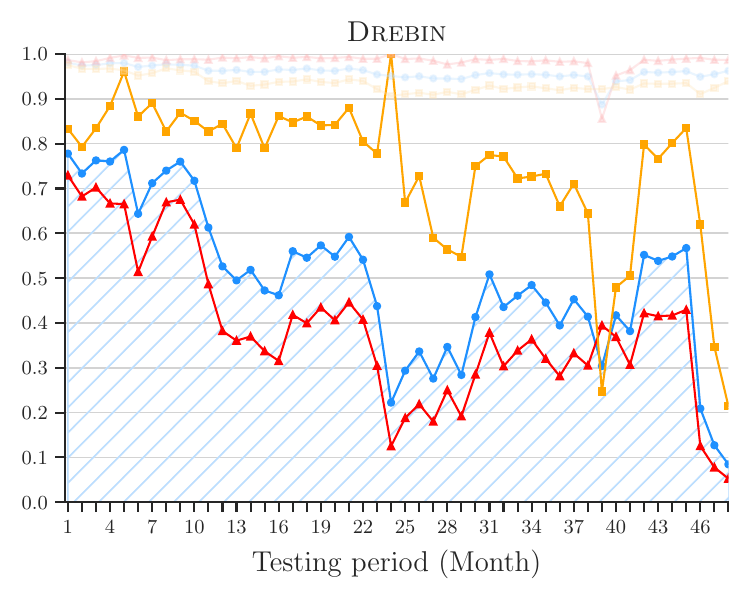}
		\caption{Top-10k features}
        \label{fig:reduce-drebin-10k}
	\end{subfigure}
	\begin{subfigure}{0.23\textwidth}
		\centering
		\includegraphics[width=\linewidth]{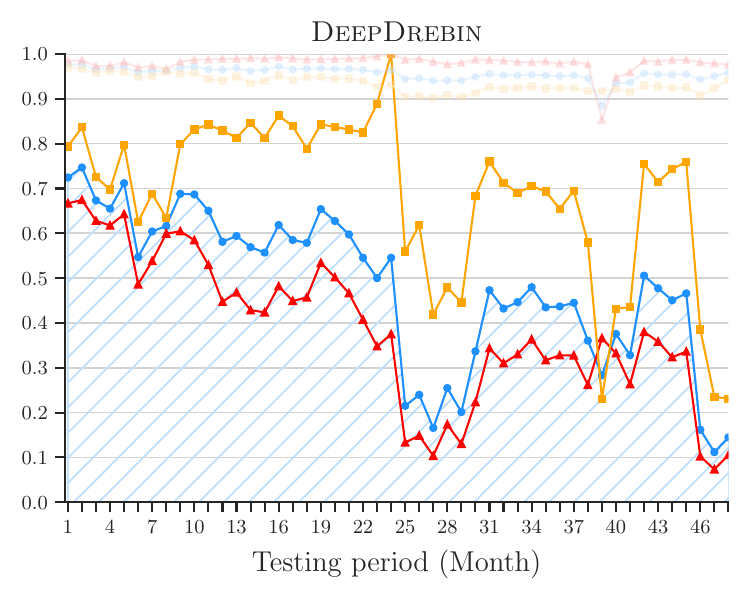}
		\caption{All features}
        \label{fig:reduce-dl-full}
	\end{subfigure}
	\begin{subfigure}{0.23\textwidth}
		\centering
		\includegraphics[width=\linewidth]{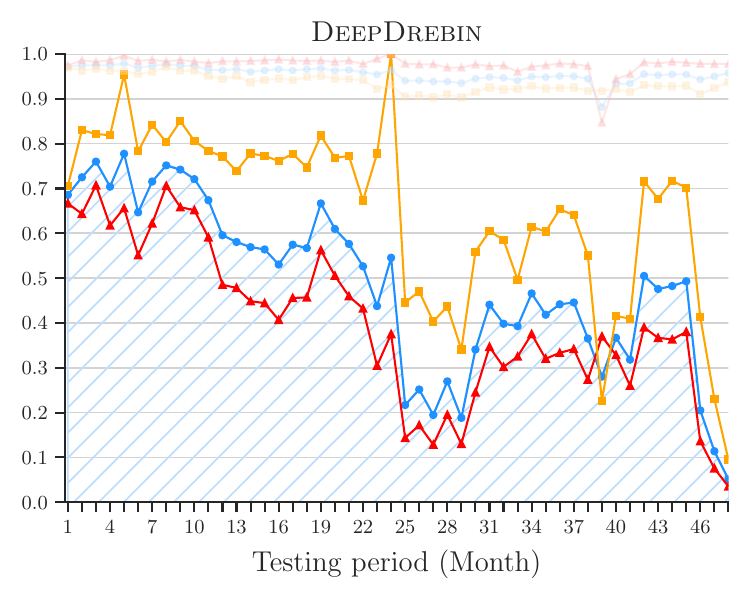}
		\caption{Top-10k features}
        \label{fig:reduce-dl-10k}
	\end{subfigure}
	\caption{Performance before and after feature selection. \textit{'All features'} stands for the approach's performance on the full extracted \drebin feature space, and \textit{'Top-10k features'} represents the performance with the most important 10,000 features which selected based on the weight vector of classifier.}
	\label{fig:compare}
\end{figure}

We reduce the \drebin feature space by selecting the top 10,000 features based on the largest absolute values in $\mathbf{w}$ \cite{demontis2017yes}. Subsequently, only the indices of these chosen features are retained for ongoing experiments. We validate the performance of both the \drebin and \dl classifiers post this feature selection. Testing results spanning from 2015 to 2018 are depicted in \autoref{fig:compare}. The left two plots, \autoref{fig:reduce-drebin-full} and \autoref{fig:reduce-drebin-10k}, illustrate the performance between \drebin's full feature space and its reduced 10,000 feature space, while the \autoref{fig:reduce-dl-full} and \autoref{fig:reduce-dl-10k} on the right are dedicated to \dl. It is important to note that after the initial model training with 2014 data, no further retraining occurs during the testing phase. Key metrics like the $F_1$-score, Precision, and Recall for both malware and goodware classes are evaluated. The results show that in both the \drebin and \dl set, the performance in the reduced feature space remains largely consistent, aligning with findings from previous studies \cite{demontis2017yes, Roy:ExpML, chen2023continuous}.

In summary, feature selection has been effective in addressing the computational and time-consuming challenges associated with large feature spaces, especially for deep learning approaches \cite{Goodfellow:DL}. Our experiments demonstrate that reducing the \drebin feature space does not negate the occurrence of concept drift; in fact, it can enhance performance by eliminating irrelevant information. Therefore, we believe that utilizing feature selection is a practical way to improve the efficiency and effectiveness of malware classifiers. Thus, we decide to use the reduced \drebin feature space, with the most important 10,000 features, in the subsequent experiments.

\section{Sources of Experimental Bias}
\label{sec:bias}

In this section, we provide the rationale for our exploration of experimental biases via conducting experiments with \drebin and \dl (\S~\ref{subsec:example}). Subsequently, we delve into the origins of temporal bias (\S~\ref{subsec:temporal}) and spatial bias (\S~\ref{subsec:spatial}) that impact ML-based classification of Android malware.

\subsection{Motivational Example}
\label{subsec:example}
\begin{table}[t]
        \centering
        \caption{$F_1$-Score results that show impact of spatial (in columns) and temporal (in rows) experimental bias. Values with red backgrounds are experimental results of \emph{unrealistic} settings similar to those considered in papers of \drebin{}~\cite{Arp:Drebin} and \dl{}~\cite{Papernot:ESORICS}; cells with green background are results in the \emph{realistic} settings we identify. The dataset consists of five years (\autoref{fig:dataset}), and each square in the sample dates column of the table represents a twelve month time-frame data: if training (resp. testing) objects are sampled from that time frame, we use a black square ($\blacksquare$); if not, we use a gray square ($\textcolor{lightgray}{\blacksquare}$).}

        \resizebox{0.9\textwidth}{!}{%

        \begin{tabular}{|r||c|c||c|c|c|c||c|c|c|c|}
			\hhline{~~~--------}

             \multicolumn{3}{c||}{} & \multicolumn{8}{c|}{{\bf
\cellcolor{LightCyan} \% mw in testing set Ts}}\\
			\hhline{~~~--------}

             \multicolumn{1}{c}{} & \multicolumn{1}{c}{} & & \multicolumn{4}{c||}{10\% (\emph{realistic})}  & \multicolumn{4}{c|}{90\% (\emph{unrealistic})} \\

			\hhline{~~~--------}

            \multicolumn{1}{c}{} & \multicolumn{1}{c}{} & & \multicolumn{4}{c||}{\bf \cellcolor{LightCyan} \% mw in training set Tr} & \multicolumn{4}{c|}{\cellcolor{LightCyan} \bf \% mw in training set Tr} \\

			\hhline{~----------}
            \multicolumn{1}{c}{} &  \multicolumn{2}{|c||}{\cellcolor{LightCyan}  {\bf Sample dates}}  & 10\% & 90\% & 10\% & 90\% & 10\% & 90\% & 10\% & 90\% \\

			\hline
            \cellcolor{LightCyan} {\bf Experimental setting} &  \multirow{1}{*}{Training} &  \multirow{1}{*}{Testing}  & \multicolumn{2}{c|}{\cellcolor{DarkCyan} \drebin{}~\cite{Arp:Drebin}} & \multicolumn{2}{c||}{\cellcolor{DarkCyan}  \dl{}~\cite{Papernot:ESORICS}} & \multicolumn{2}{c|}{\cellcolor{DarkCyan}  \drebin{}~\cite{Arp:Drebin}} & \multicolumn{2}{c|}{\cellcolor{DarkCyan} \dl{}~\cite{Papernot:ESORICS}} \\
		\hline \hline

            \multirow{2}{*}{{\bf 10-fold CV}} & {\tt gw}:  $\blacksquare \blacksquare \blacksquare \blacksquare \blacksquare$  & {\tt gw}:  $\blacksquare \blacksquare \blacksquare \blacksquare \blacksquare $ &  \cellcolor{LightRed}  & \multirow{2}{*}{0.55} & \cellcolor{LightRed}  & \multirow{2}{*}{0.51} & \multirow{2}{*}{0.91} & \multirow{2}{*}{0.98} & \multirow{2}{*}{0.88} &    \\
            & {\tt mw}: $\blacksquare \blacksquare \blacksquare \blacksquare \blacksquare$  & {\tt mw}: $\blacksquare \blacksquare \blacksquare \blacksquare \blacksquare $  &  \cellcolor{LightRed} \multirow{-2}{*}{\bf 0.85} & & \cellcolor{LightRed} \multirow{-2}{*}{\bf 0.84}& & & & &  \multirow{-2}{*} {0.98}  \\
                    \hline

                \multirow{2}{*}{{\bf Temporally inconsistent}} & {\tt gw}: $\textcolor{lightgray}{\blacksquare}  \textcolor{lightgray}{\blacksquare} \blacksquare \textcolor{lightgray}{\blacksquare}  \textcolor{lightgray}{\blacksquare}$  & {\tt gw}: $\blacksquare \blacksquare \textcolor{lightgray}{\blacksquare} \blacksquare \blacksquare$  & \multirow{2}{*}{0.64}  & \multirow{2}{*}{0.48} & \multirow{2}{*}{0.62} & \multirow{2}{*}{0.38} & \multirow{2}{*}{0.71} & \multirow{2}{*}{0.49} & \multirow{2}{*}{0.66} &   \\
                     & {\tt mw}: $\textcolor{lightgray}{\blacksquare}  \textcolor{lightgray}{\blacksquare} \blacksquare \textcolor{lightgray}{\blacksquare}  \textcolor{lightgray}{\blacksquare}$  & {\tt mw}: $\blacksquare \blacksquare \textcolor{lightgray}{\blacksquare} \blacksquare \blacksquare$  &  & &  & & & & & \multirow{-2}{*}{0.92} \\
            \hline

             \multirow{1}{*}{{\bf Temporally inconsistent}}& {\tt gw}: $\textcolor{lightgray}{\blacksquare} \blacksquare  \textcolor{lightgray}{\blacksquare}
             \textcolor{lightgray}{\blacksquare}
             \textcolor{lightgray}{\blacksquare}$  & {\tt gw}: $\textcolor{lightgray}{\blacksquare} \textcolor{lightgray}{\blacksquare} \textcolor{lightgray}{\blacksquare} \textcolor{lightgray}{\blacksquare} \blacksquare $  &   
             \multirow{2}{*}{0.30}& \multirow{2}{*}{0.42} & \multirow{2}{*}{0.37} &  \multirow{2}{*}{0.42}  & 
             
             \multirow{2}{*}{0.31} & \multirow{2}{*}{0.77} & \multirow{2}{*}{0.39} &  \\
            {\bf gw/mw windows} & {\tt mw}: $\blacksquare  \textcolor{lightgray}{\blacksquare} \textcolor{lightgray}{\blacksquare} \textcolor{lightgray}{\blacksquare}  \textcolor{lightgray}{\blacksquare}$  & {\tt mw}: $\textcolor{lightgray}{\blacksquare} \textcolor{lightgray}{\blacksquare} \blacksquare \textcolor{lightgray}{\blacksquare} \textcolor{lightgray}{\blacksquare} $  &  & &  & & & & &   \multirow{-2}{*}{0.78}\\
                        \hline

            \multirow{1}{*}{\bf Temporally consistent}& {\tt gw}: $\blacksquare \textcolor{lightgray}{\blacksquare} \textcolor{lightgray}{\blacksquare} \textcolor{lightgray}{\blacksquare}  \textcolor{lightgray}{\blacksquare}$  & {\tt gw}: $\textcolor{lightgray}{\blacksquare} \blacksquare \blacksquare \blacksquare \blacksquare $
            & \cellcolor{LightGreen}  & \cellcolor{LightGreen} & \cellcolor{LightGreen}  & \cellcolor{LightGreen} & \multirow{2}{*}{0.53} & \multirow{2}{*}{0.80} & \multirow{2}{*}{0.39} & \multirow{2}{*}{0.84} \\
                				  ({\it realistic}) & {\tt mw}: $\blacksquare \textcolor{lightgray}{\blacksquare} \textcolor{lightgray}{\blacksquare} \textcolor{lightgray}{\blacksquare}  \textcolor{lightgray}{\blacksquare}$  & {\tt mw}: $ \textcolor{lightgray}{\blacksquare} \blacksquare \blacksquare \blacksquare \blacksquare $  & \multirow{-2}{*}{\cellcolor{LightGreen} {\color{black} \bf 0.47}} & \multirow{-2}{*}{\cellcolor{LightGreen} { \color{black} \bf 0.39}}  &  \multirow{-2}{*}{\cellcolor{LightGreen}{\color{black}\bf 0.36}}   &  \multirow{-2}{*}{\cellcolor{LightGreen} {  \color{black} \bf 0.39}} & & & & \\
            \hline

        \end{tabular}
        }
        \vspace{1em}

        \label{tab:bias}

    \end{table}

We present a motivating example where we manipulate experimental biases to better illustrate the problem. The results are summarized in \autoref{tab:bias}, which showcases the $F1$-score for \drebin and \dl under different experimental configurations (as shown in cells under the \textit{Experimental settings} column). The rows represent various sources of temporal experimental bias, while the columns represent different sources of spatial experimental bias.

In the \textit{Sample dates} column of \autoref{tab:bias}, squares (${\blacksquare}$ / ${\textcolor{lightgray}\blacksquare}$) indicate the time frames from which the training and testing samples are obtained. Each square corresponds to a twelve-month period within the range from January 2014 to December 2018. Black squares ($\blacksquare$) indicate that samples are taken from that specific time frame, while gray squares ($\textcolor{lightgray}\blacksquare$) represent periods that are not utilized. On the right side of the table, the columns denote different percentages of malware in the training set ($\mathit{Tr}$) and the testing set ($\mathit{Ts}$).

The data presented in \autoref{tab:bias} reveal that both \drebin and \dl exhibit significantly lower performance in realistic scenarios (highlighted in bold with a green background in the last row, corresponding to columns with 10\% malware in testing) compared to settings similar to those described in prior works such as \cite{Arp:Drebin, Papernot:ESORICS} (highlighted in bold with a red background). This discrepancy can be attributed to unintentional experimental biases, which we elaborate on in the following sections.

\begin{tcolorbox}

\textbf{Note.} We would like to clarify the reference to the similar settings in \cite{Arp:Drebin, Papernot:ESORICS} mentioned in the cells with a red background in Table \ref{tab:bias}. The original paper of \drebin \cite{Arp:Drebin} utilizes a hold-out approach by performing 10 random splits, with 66\% of the data used for training and 33\% for testing. On the other hand, the paper on \dl \cite{Papernot:ESORICS} briefly mentions using the same dataset and experimental setting as \cite{Arp:Drebin}. Since hold-out is nearly equivalent to k-fold cross-validation (CV) and is susceptible to the same spatio-temporal biases, for the sake of simplicity in this section, we refer to a k-fold cross validation setting for both \drebin and \dl.

\end{tcolorbox}

\subsection{Temporal Experimental Bias}
\label{subsec:temporal}

\emph{Concept drift} is a problem that occurs in machine learning when a model becomes obsolete as the distribution of incoming data at test-time differs from that of training data, i.e., when the assumption does not hold that data is independent and identically distributed (i.i.d.)~\cite{Jordaney:Transcend, barbero2022transcending}. In the ML community, this problem is also known as \emph{dataset shift}~\cite{DatasetShift}. \emph{Time decay} is the decrease in model performance over time caused by dataset shift.

Concept drift in malware combined with similarities among malware within the same family causes \emph{k-fold cross validation} (CV) to be \emph{positively biased}, artificially inflating the performance of malware classifiers~\cite{Allix:Timeline,Miller:Reviewer,Miller:Thesis}. 
K-fold CV is likely to include in the training set at least one sample of each malware family in the dataset, whereas new families will be unknown at training time in a real-world deployment. The all-black squares in \autoref{tab:bias} for 10-fold CV refer to each training/testing fold of the 10 iterations containing at least one sample from each time frame. 
The use of k-fold CV is widespread in malware classification research~\cite{Miller:Thesis,Perdisci:URLs,Suarez:DroidSieve,Maggi:URLs,Dahl:Large,Tahan:Mal,Markel:Building,Laskov:PDF,Zhang:Semantics,Curtsinger:Zozzle}; while a useful mechanism to prevent overfitting~\cite{Bishop:ML} or estimate the performance of a classifier in the \emph{absence} of concept drift when the i.i.d. assumption holds (see considerations in \S~\ref{subsec:reveal performance}), it has been unclear how it affects the real-world performance of machine learning techniques with non-stationary data that are affected by time decay. Hence, in the first row of \autoref{tab:bias}, we quantify the performance impact with 10-fold CV in the Android domain.

The second row of \autoref{tab:bias} reports an experiment in which a classifier’s ability to detect both past and future objects is evaluated \cite{Arp:Drebin, Papernot:ESORICS}. Although this characteristic is important, high performance should be expected from a classifier in such a scenario: if the classifier contains at least one variant of a past malware family, it will likely identify similar variants. We thus believe that experiments on the performance achieved on the detection of past malware can be misleading; therefore future research should focus on building malware classifiers that are robust against time decay.

In the third row of \autoref{tab:bias}, we identify a novel form of temporal bias that arises when goodware and malware correspond to different time periods, often due to originating from distinct data sources (e.g., as observed in \cite{Mariconti:MaMaDroid}). The black and gray squares in \autoref{tab:bias} indicate that, even though the testing objects for malware come after the training objects, the time windows for goodware and malware do not overlap. This situation can lead to the classifier learning to distinguish between applications from different time periods rather than differentiating between goodware and malware, thereby resulting in artificially inaccurate performance. For instance, spurious features such as new API methods may exhibit strong discriminatory power simply because malicious applications predate the introduction of those APIs.

Temporal inconsistency in data can result in \textit{Spurious Correlations}. As described by \cite{Arp:DoDont}, this refers to a scenario where a ML model forms shortcut patterns to differentiate classes based on artifacts that are irrelevant to the actual security issue. For instance, consider a malware detector that is trained on a dataset heavily dominated by certain malware families prevalent at the time. The model might inadvertently learn to identify features that are specific to these families rather than the broader characteristics of malware. Such features, although effective in the short-term, may become redundant as new malware families emerge, which were not present during the training phase. This leads to a temporal inconsistency in the model's performance, as the model is tailored to the specific context of the training data and may not generalize well to future threats \cite{Arp:DoDont}. 
Consequently, it is essential for designers to be mindful of and avoid potential temporal inconsistencies during the data collection and sampling process.

The last row of \autoref{tab:bias} shows that the realistic setting, where
training is temporally precedent to testing, causes the worst classifier
performance in the majority of cases. We present decay plots and a more detailed discussion in \S~\ref{sec:Space-Time Aware Evaluation}.

\subsection{Spatial Experimental Bias}
\label{subsec:spatial}

We identify two main types of spatial experimental bias based on assumptions on percentages of malware in testing and training sets. For all experiments in this section, we assume temporal consistency, meaning that the model is trained on data from 2014 and tested on data from 2015 onward (as the setting indicated in the last row of \autoref{tab:bias}). This temporal setup allows us to examine spatial bias independently, without the confounding influence of temporal bias.

\begin{figure}[t]
\centering
	\begin{subfigure}{0.24\textwidth}
		\centering
		\includegraphics[width=\linewidth]{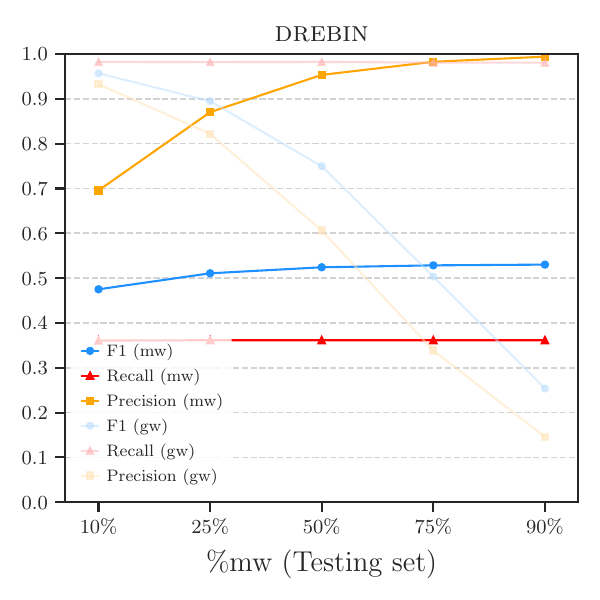}
		\caption{Train: 10\% mw}
	\end{subfigure}
	\begin{subfigure}{0.24\textwidth}
		\centering
		\includegraphics[width=\linewidth]{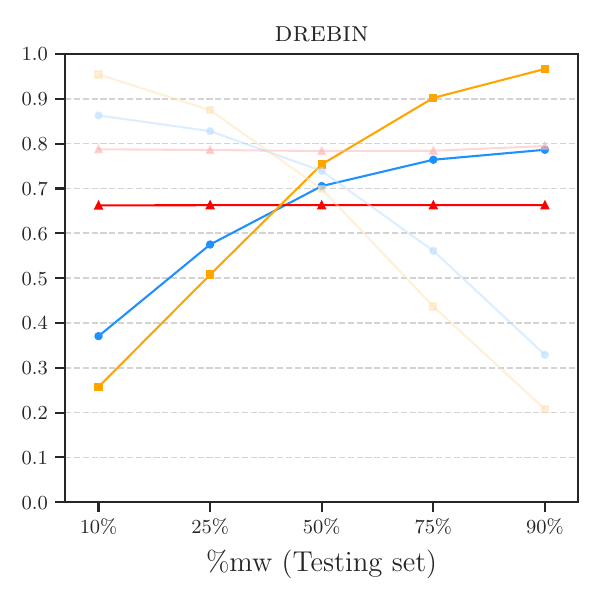}
		\caption{Train: 90\% mw}
	\end{subfigure}
	\begin{subfigure}{0.24\textwidth}
		\centering
		\includegraphics[width=\linewidth]{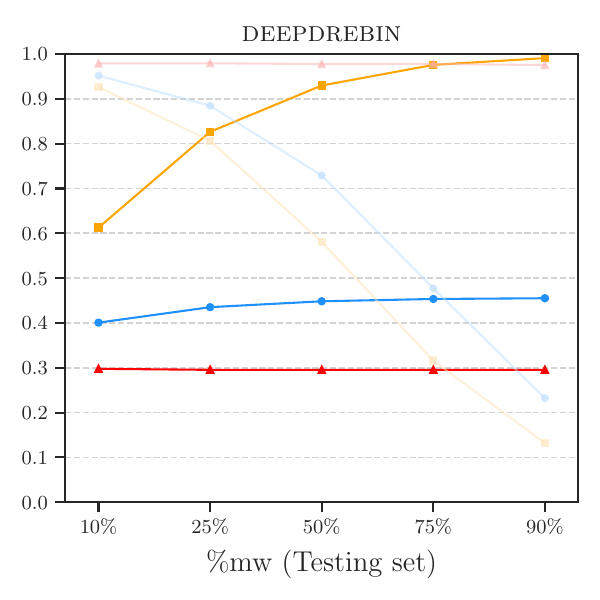}
		\caption{Train: 10\% mw}
	\end{subfigure}
	\begin{subfigure}{0.24\textwidth}
		\centering
		\includegraphics[width=\linewidth]{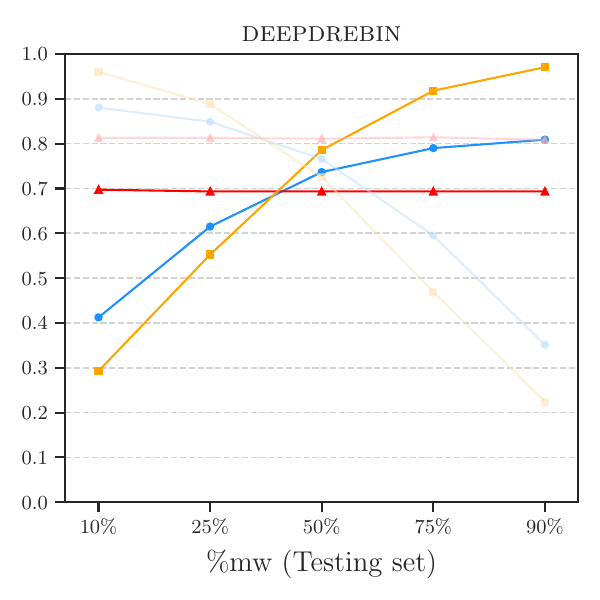}
		\caption{Train: 90\% mw}
	\end{subfigure}
	\caption{{\it Spatial experimental bias in testing}. The models are trained on data from 2014 and tested on data from the remaining four years. In this unrealistic setting, where the percentage of malware in the testing is artificially increased, Precision for malware increases while Recall remains similar. Consequently, the overall $F_1$-Score also increases with the rising percentage of malware in the testing. However, it is important to note that this setting with more malware than goodware in testing does not reflect the true in-the-wild distribution of 10\% malware (\S~\ref{subsec:Malware Ratio}), rendering it unrealistic and leading to biased results.}
	\label{fig:space-test-bias}
\end{figure}

\textbf{Spatial experimental bias in testing.} The percentage of malware in the testing distribution needs to be estimated (\S~\ref{subsec:Malware Ratio}) and \textit{cannot} be changed, if one wants results to be representative of in-the-wild deployment of the malware classifier. To understand why this leads to biased results, we artificially vary the testing distribution to illustrate our point. \autoref{fig:space-test-bias} reports performance ($F_1$-Score, Precision, Recall) for increasing the percentage of malware during testing on the x-axis. We change the percentage of malware in the testing set by randomly down-sampling goodware, so the number of malware remains fixed throughout the experiments. For completeness, we report the two training settings from \autoref{tab:bias} with 10\% and 90\% malware, respectively. 

Note that we choose to down-sample goodware (gw) to achieve up to 90\% of malware (mw) for testing because of the computational and storage resources required to achieve such a ratio by oversampling. However, this does not alter the conclusions of our analysis. Consider a scenario where gw quantity is constant, but mw is increased via oversampling. The precision ($P_{mw} = TP/(TP+FP)$) would increase because TPs would increase for any mw detection, and FPs would not change—because the number of gw remains the same; if training (resp. testing) observations are sampled from a distribution similar to the mw in the original dataset (e.g., new training mw is from 2014 and new testing mw comes from 2015 and 2016), then Recall ($R_{mw} = TP/(TP+FN)$) would be stable—it would have the same proportions of TPs and FNs because the classifier will have a similar predictive capability for finding mw. Hence, if the number of mw in the dataset increases, the $F_1$-Score would increase as well, because Precision increases while Recall remains stable.

Let us first focus on the malware performance (highlighted lines). All plots in \autoref{fig:space-test-bias} exhibit consistent Recall, and increasing Precision when increasing percentage of malware in the testing. Precision for the malware (mw) class - the positive class - is defined as $P_{mw} = TP/(TP+FP)$ and Recall as $R_{mw} = TP/(TP+FN)$. 
In this scenario, we can observe that TPs (i.e., malware objects correctly classiﬁed as malware) and FNs (i.e., malware objects incorrectly classiﬁed as goodware) do not change, because the number of malware does not increase; hence, Recall remains stable. The increase in number of FPs (i.e., goodware objects misclassified as malware) decreases as we reduce the number of goodware in the dataset; hence, Precision improves. Since the $F_1$-Score is the harmonic mean of Precision and Recall, it goes up with Precision. We also observe that, inversely, the Precision for the goodware (gw) class - the negative class - $P_{gw} = TN/(TN +FN)$ decreases (see yellow dimmed lines in \autoref{fig:space-test-bias}), because we are reducing the TNs while the FNs do not change. This example shows how considering an unrealistic testing distribution with more malware than goodware in this context (\S~\ref{subsec:Malware Ratio}) positively inflates Precision and hence the $F_1$-Score of classifiers.

\begin{figure}[]
\centering
	\begin{subfigure}{0.31\textwidth}
		\centering
		\includegraphics[width=\linewidth]{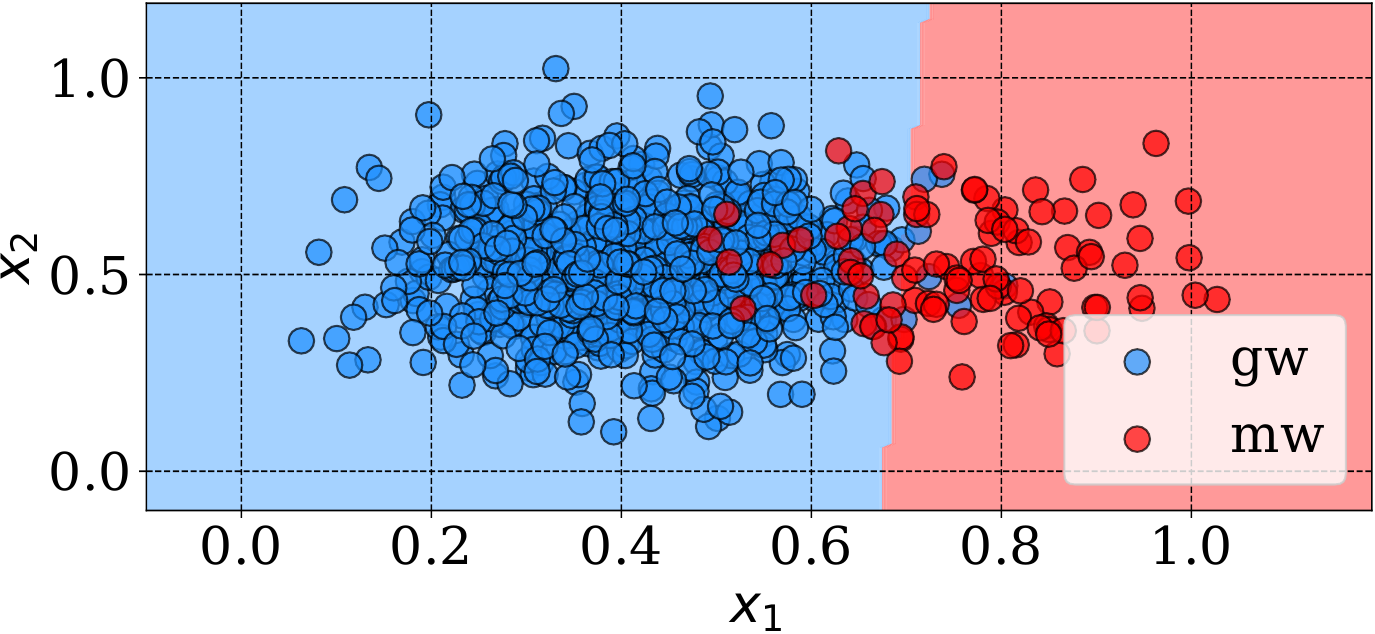}
		\caption{Train: 10\% mw; Test: 10\% mw}
	\end{subfigure}
	\begin{subfigure}{0.31\textwidth}
		\centering
		\includegraphics[width=\linewidth]{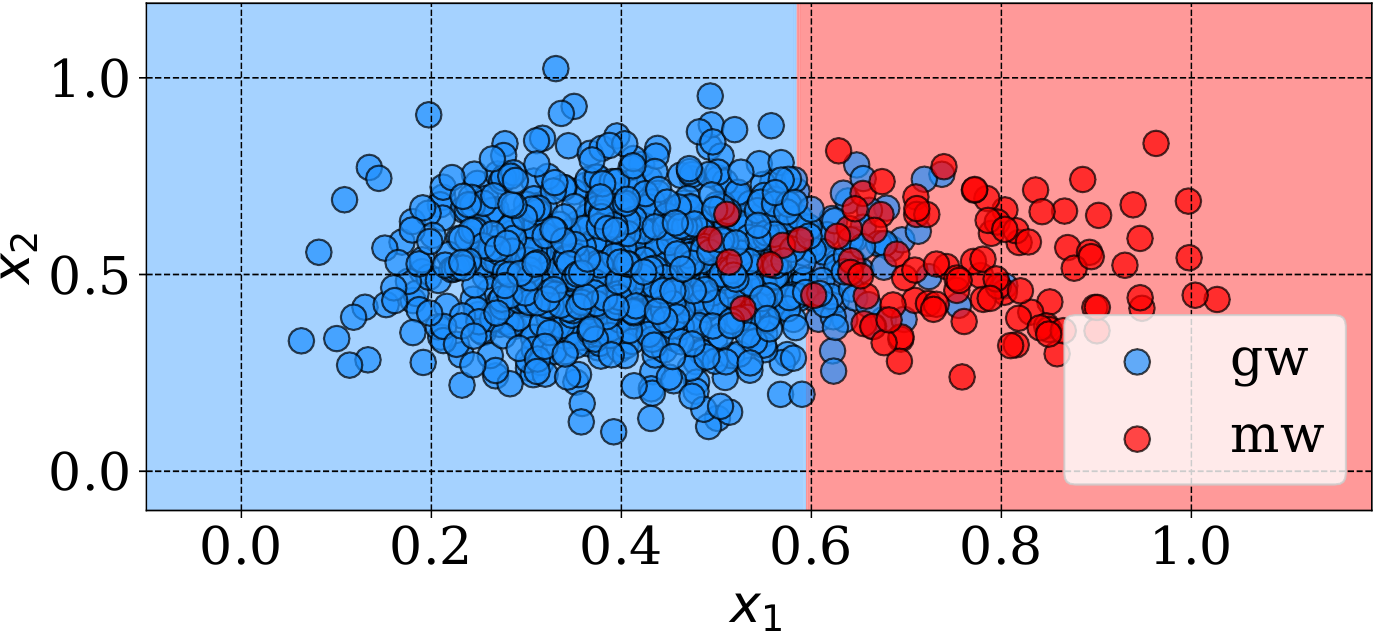}
		\caption{Train: 50\% mw; Test: 10\% mw}
	\end{subfigure}
	\begin{subfigure}{0.31\textwidth}
		\centering
		\includegraphics[width=\linewidth]{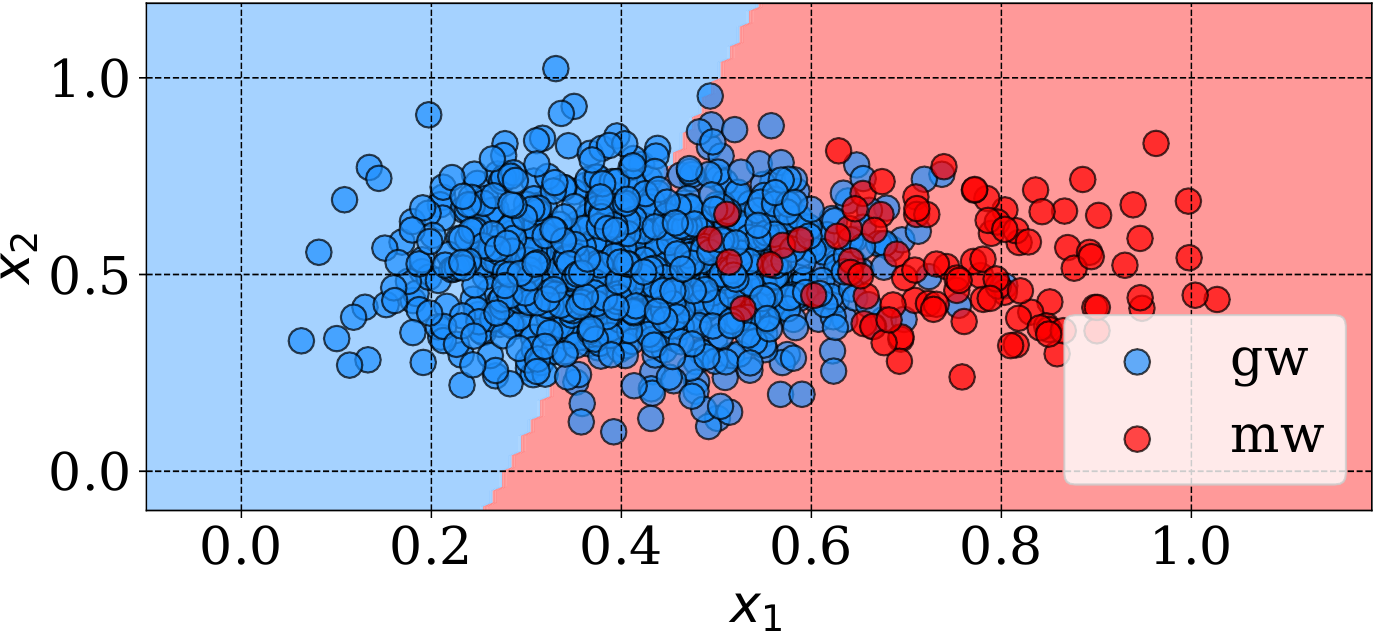}
		\caption{Train: 90\% mw; Test: 10\% mw}
	\end{subfigure}
	
	\caption{Motivating example for the intuition of \emph{spatial experimental bias in training} with Linear-SVM and two features, $x_1$ and $x_2$. The training changes, but the testing points are fixed: 90\% gw and 10\% mw. When the percentage of malware in the training increases, the decision boundary moves towards the goodware class, improving Recall for malware but decreasing Precision.}
	\label{fig:toysvm}
\end{figure}

\begin{figure}[t]
	\centering
	\begin{subfigure}{0.23\textwidth}
		\centering
		\includegraphics[width=\linewidth]{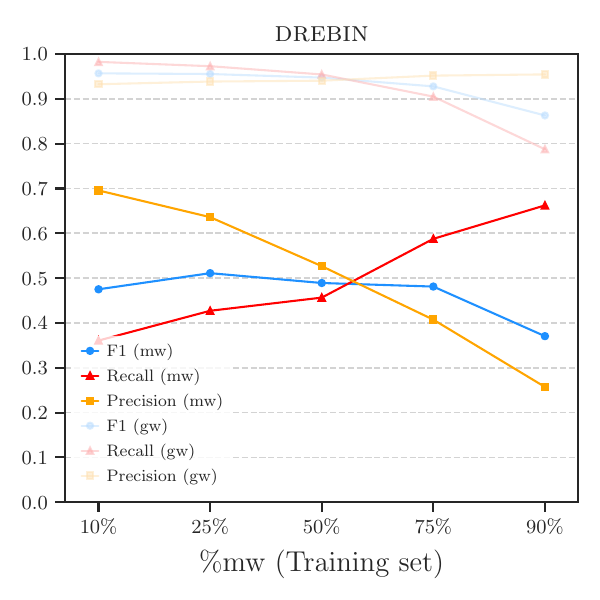}
		\caption{Test: 10\% mw}
	\end{subfigure}
 \hfill
	\begin{subfigure}{0.23\textwidth}
		\centering
		\includegraphics[width=\linewidth]{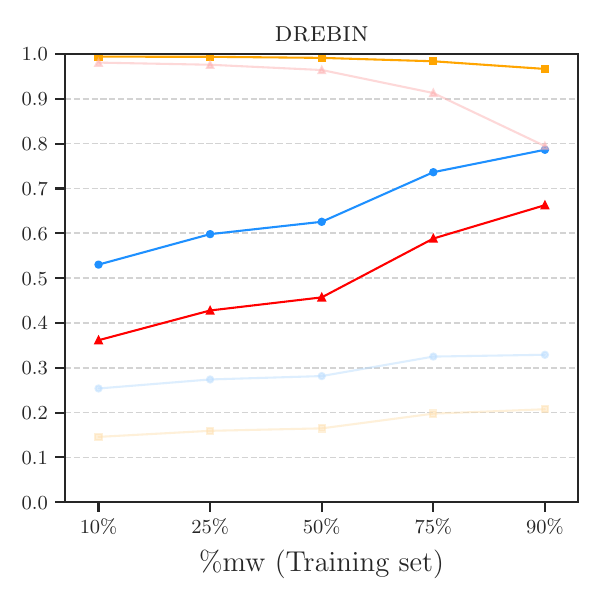}
		\caption{Test: 90\% mw}
	\end{subfigure}
  \hfill
	\begin{subfigure}{0.23\textwidth}
		\centering
		\includegraphics[width=\linewidth]{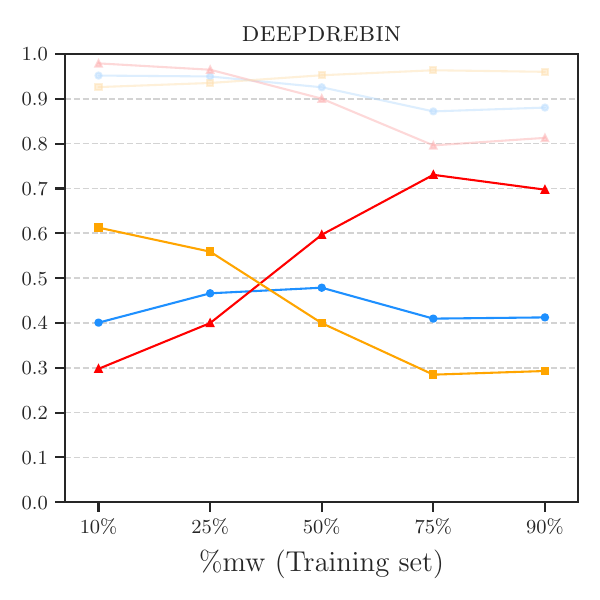}
		\caption{Test: 10\% mw}
	\end{subfigure}
  \hfill
	\begin{subfigure}{0.23\textwidth}
		\centering
		\includegraphics[width=\linewidth]{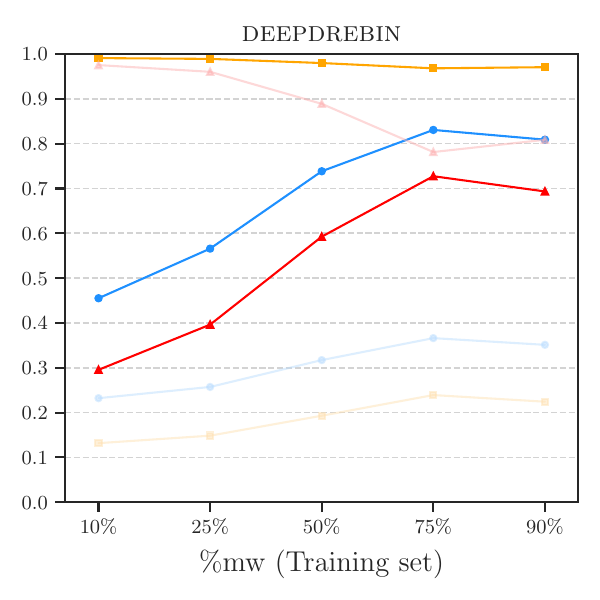}
		\caption{Test: 90\% mw}
	\end{subfigure}
	\caption{{\it Spatial experimental bias in training.} The models were trained on data from 2014 and tested on the remaining four years data. As the percentage of malware in the training set increases, Precision decreases while Recall increases, aligning with the motivations illustrated in the example of \autoref{fig:toysvm}. In \S~\ref{subsec:tuning alg}, we present an algorithm to determine the optimal training configuration for optimizing Precision, Recall, or $F_1$-Score based on user requirements.}
	\label{fig:space-train-bias}
\end{figure}

\textbf{Spatial experimental bias in training.} The impact of altering the malware-to-goodware ratios in training can be understood through a motivating example with a linear SVM in a 2D feature space, using features $x_1$ and $x_2$. In this example, \autoref{fig:toysvm} presents three scenarios, each with the same 10\% malware in testing but varying percentages of malware in training: 10\%, 50\%, and 90\%.

It is observed that as the percentage of malware in training increases, the hyperplane of the SVM shifts towards the goodware region. Specifically, it improves the Recall of malware while reducing its Precision. Conversely, the opposite trend is observed for goodware. In order to minimize the overall error rate $Err = (FP+FN)/(TP+TN+FP+FN)$ (i.e., maximize Accuracy), it is advisable to train the dataset with the same distribution that is expected in the testing. However, in certain cases, there may be a greater interest in identifying objects of the minority class (e.g., "more malicious samples") by improving Recall while ensuring a constraint on the maximum FPR.

\autoref{fig:space-train-bias} shows the performance for \drebin and \dl, with increasing percentages of malware in training on the x-axis. For the sake of completeness (as it is not possible to artificially change the test distribution for realistic evaluations), results are reported for both 10\% malware in testing and 90\% malware in testing. However, it should be noted that in the Android setting, the estimated proportion of malware in the wild is 10\% (see \S~\ref{subsec:Malware Ratio}). These plots confirm the trends observed in our motivating example (\autoref{fig:toysvm}), where $R_{mw}$ increases while $P_{mw}$ decreases. For the plots with 10\% malware in testing, it can be observed that there is a point at which the $F_1$-Score for malware is maximized while keeping the error for the goodware (gw) class within 5\%.

In \S~\ref{subsec:tuning alg}, we propose a novel algorithm to enhance the performance of the malware class based on the user's objective (e.g., high Precision, Recall, or $F_1$-Score), while ensuring a maximum tolerated error. Furthermore, in \S~\ref{sec:Space-Time Aware Evaluation}, we introduce constraints and metrics to ensure bias-free evaluations and reveal counter-intuitive results.

\section{Space-Time Aware Evaluation}
\label{sec:Space-Time Aware Evaluation}

We now outline the process for conducting an evaluation of an Android malware classifier that is free from spatio-temporal bias. Firstly, we establish a novel set of constraints that must be adhered to in order to ensure realistic evaluations (\S~\ref{subsec:constrains}). Secondly, we introduce a new time-aware metric called AUT, which provides a comprehensive measure of the impact of time decay on a classifier (\S~\ref{subsec:performance metrics}). Thirdly, we propose a novel tuning algorithm that empirically optimizes the performance of a classifier, while considering a maximum acceptable error (\S~\ref{subsec:tuning alg}). Finally, we introduce \tesseract, a framework that enables unbiased evaluations and presents counter-intuitive results (\S~\ref{subsec:reveal performance}). For ease of reference, we provide a table of major symbols used throughout the remainder of this paper in Appendix \ref{app:symbol} to enhance readability.

\subsection{Evaluation Constraints}
\label{subsec:constrains}

Let us consider $\mathit{D}$ as a labeled dataset with two classes: malware (positive class) and goodware (negative class). Let us define $s_i \in \mathit{D}$ as an \emph{object} (e.g., testing sample) with timestamp $\mathit{time}(s_i)$. To evaluate the classifier, the dataset $\mathit{D}$ must be split into a training dataset $\mathit{Tr}$ with a time window of size $W$, and a testing dataset $\mathit{Ts}$ with a time window of size $S$.
Here, we consider $S > W$ in order to estimate long-term performance and robustness to decay of the classifier. A user may consider different time splits depending on their objectives, provided each split has a significant number of samples. We emphasize that, although we have the labels of objects in $\mathit{Ts} \subseteq $ $\mathit{D}$, all the evaluations and tuning algorithms \emph{must} assume that labels $y_i$ of objects $s_i \in \mathit{Ts}$ are unknown.

To evaluate performance over time, the test set $\mathit{Ts}$ must be split into time-slots of size $\Delta$. For example, for a testing set time window of size $S = 2 \:\, \mathrm{years}$, we may have $\Delta = 1 \:\, \mathrm{month}$, for time window of size $S = 6 \:\, \mathrm{months}$, we may choose $\Delta = 1 \:\, \mathrm{week}$, etc. This granularity parameter is chosen by the user, but it is important that the chosen $\Delta$ allows for a statistically significant number of objects in each test window $[t_i, t_i+ \Delta)$.

We now formalize three constraints that must be enforced when dividing
$\mathit{D}$ into $\mathit{Tr}$ and $\mathit{Ts}$ for a realistic
setting that avoids spatio-temporal experimental bias (\S~\ref{sec:bias}).
While C1 was proposed in past work~\cite{Allix:Timeline, Miller:Reviewer},  we are the first to propose C2 and C3---which we show to be fundamental in \S~\ref{subsec:reveal performance}.

\vspace{1em}

{\bf (C1) Temporal training consistency.} All the
objects in the training must be {\it strictly} temporally
precedent to the testing~ones:
\begin{equation}
	\mathit{time}(s_i) < \mathit{time}(s_j), \forall s_i \in \mathit{Tr} , \forall s_j \in \mathit{Ts}
	\label{eq:c1}
\end{equation}
where $s_i$ (resp. $s_j$) is an object in the training set $\mathit{Tr}$
(resp. testing set $\mathit{Ts}$). Eq.~\ref{eq:c1} must hold; its
violation inflates the results by including
future knowledge in the classifier (\S~\ref{subsec:temporal}).

\vspace{1em}

{\bf (C2) Temporal gw/mw time-window consistency.} In every
testing slot of size $\Delta$, all test objects $s_i$ must be from the same
time window: 
\begin{equation}
	t^{\mathit{min}}_i \leq \mathit{time}(s_i) \leq t^{\mathit{max}}_i,\quad \forall s_i \text{ in time slot } [t_i, t_i+\Delta)
	\label{eq:c2}
\end{equation}

where $t^{\mathit{min}}_i = \min_k \mathit{time}(s_i)$ and
$t^{\mathit{max}}_i = \max_k \mathit{time}(s_i)$. The same should hold for the training: although violating \autoref{eq:c2} in the training data does not bias the evaluation, it may affect the sensitivity of the classifier to unrelated artifacts.
\autoref{eq:c2} has been violated in the past when goodware and malware have been collected from different time windows (e.g., \cite{Mariconti:MaMaDroid})---if violated, the results are biased because the classifier may learn and test on artifactual behaviors that, for example, distinguish goodware from malware just by their different API versions.

\vspace{1em}

{\bf (C3) Realistic malware-to-goodware ratio in testing.}
Let us define~$\varphi$ as the average percentage of malware in training data, and $\delta$ as the average percentage of malware in the testing data. Let $\hat{\sigma}$ be the estimated percentage of malware in the wild. To have a realistic evaluation, the average percentage of malware in the testing ($\delta$) must be as close as possible to the estimated percentage of malware in the wild ($\hat{\sigma}$), so that:
\begin{equation}
	\delta \simeq \hat{\sigma}
	\label{eq:c3}
\end{equation}

For example, we have estimated that in the Android scenario, especially during our evaluated period (2014 - 2018), goodware is predominant over malware, with $\hat{\sigma} \approx 10\%$ (\S~\ref{subsec:Malware Ratio}). If C3 is violated by overestimating the percentage of malware, the results are positively inflated (\S~\ref{subsec:spatial}).
We highlight that, although the testing distribution $\delta$ cannot be changed (in order to get realistic results), the percentage of malware in the training $\varphi$ may be tuned (\S~\ref{subsec:tuning alg}).

\subsection{Time-aware Performance Metrics}
\label{subsec:performance metrics}

We introduce a time-aware performance metric that allows for the comparison of different classifiers while considering time decay. Let $\Theta$ be a
classifier trained on~$\mathit{Tr}$; we capture the performance of $\Theta$
for each time frame $[t_i, t_i+\Delta)$ of the testing set~$\mathit{Ts}$ (e.g., each month). We identify two options to represent per-month performance:
\begin{itemize}

\item {\bf Point estimates} (\emph{pnt}): The value plotted on the $Y$-axis for $X_k = k \Delta$ (where $k$ is the test slot number) computes the performance metric (e.g., $F_1$-Score) only based on predictions $\hat{y}_i$ of $\Theta$ and true labels $y_i$ in the interval $[W+(k-1) \Delta, W + k \Delta)$.

\item {\bf Cumulative estimates} (\emph{cml}): The value plotted on the $Y$-axis for $X_k = k \Delta$ (where $k$ is the test slot number) computes the performance metric (e.g., $F_1$-Score) only based on predictions $\hat{y}_i$ of $\Theta$ and true labels $y_i$ in the cumulative interval $[W, W + k \Delta)$.

\end{itemize}

Point estimates are always to be preferred to represent the real performance of an algorithm. The cumulative estimates can be used to highlight a smoothed trend and to show overall performance up to a certain point, but can be misleading if reported on their own if objects are too sparsely distributed in some test slots $\Delta$. Hence, we report only point estimates in the remainder of the paper, while an example of cumulative estimate plots is reported in Appendix~\ref{app:cml}. 

To facilitate the comparison of different time decay plots, we define a new metric, {\it Area Under Time} ({\bf AUT}), the area under the performance curve over time. Formally, based on the trapezoidal rule (as in AUROC~\cite{Bishop:ML}), AUT is defined as follows:
\begin{equation}
\displaystyle AUT(\mathbb{P},N) = \frac{1}{N-1} \sum_{k=1}^{N-1}  \frac{[\mathbb{P}(X_{k+1}) + \mathbb{P}(X_k)]}{2} \label{eq:aut}
\end{equation}
where: $\mathbb{P}(X_k)$ is the value of the point estimate of the performance metric $\mathbb{P}$ (e.g., $F_1$) evaluated at point~$X_k := (W + k \Delta)$; $N$ is the number of test slots, and $1/(N-1)$ is a normalization factor so that AUT $\in [0,1]$.
The perfect classifier with robustness to time decay in the time
window $S$ has $\mathrm{AUT}=1$. By default, AUT is computed as the area under point estimates, as they capture the trend of the classifier over time more closely; if the AUT is computed on cumulative estimates, it should be explicitly marked as $\mathrm{AUT}_{\emph{cml}}$. As an example, $\mathrm{AUT}(F_1, 12m)$ is the point estimate of $F_1$-Score considering time decay for a period of 12 months, with a 1-month interval.

We highlight that the simplicity of computing the AUT should be seen
as a benefit rather than a drawback; it is a simple yet effective
metric that captures the performance of a classifier with respect to
time decay, de-facto promoting a fair comparison across different
approaches.

\begin{tcolorbox}
{\bf AUT}($\mathbb{P}$,$N$) is a metric that allows us to evaluate performance $\mathbb{P}$ of a malware classifier against time decay over $N$ time units in realistic experimental settings---obtained by enforcing C1, C2, and C3 (\S~\ref{subsec:constrains}). The next sections leverage AUT for tuning classifiers and comparing different solutions (\S~\ref{subsec:reveal performance}).
\end{tcolorbox}

\subsection{Tuning Training Ratio}
\label{subsec:tuning alg}

We propose a novel algorithm that allows for the adjustment of the
training ratio $\varphi$ when the dataset is imbalanced, in order to
optimize a user-specified performance metric ($F_1$-Score, Precision, or
Recall) on the minority class, subject to a maximum tolerated error, while  aiming to reduce time decay. The high-level intuition of the impact of changing $\varphi$ is described in \S~\ref{subsec:spatial}. We also observe that ML literature has shown ROC curves to be misleading on highly imbalanced datasets~\cite{davis2006relationship,He:Imbalanced}. Choosing different thresholds on ROC curves \emph{shifts} the decision boundary, but (as seen in the motivating example of \autoref{fig:toysvm}) re-training with different ratios $\varphi$ (as in our algorithm) also changes the shape of the decision boundary, better representing the minority class.

Our tuning algorithm is inspired by one proposed by Weiss and Provost~\cite{Weiss:ReBalance}; they propose a progressive sampling of training objects to collect a dataset that improves AUROC performance of the minority class in an imbalanced dataset. However, they did not take temporal constraints into account (\S~\ref{subsec:temporal}), and heuristically optimize only AUROC.
Conversely, we enforce C1, C2, C3 (\S~\ref{subsec:constrains}), and rely on AUT to achieve three possible targets for the malware class: higher $F_1$-Score, higher Precision, or higher Recall. Also, we assume that the user already has a training dataset $\mathit{Tr}$ and wants to use as many objects from it as possible, while still achieving a good performance trade-off; for this purpose, we perform a \emph{progressive sub-sampling} of the goodware class.

Algorithm~\ref{alg:varphi} formally presents our methodology for tuning the parameter $\varphi$ to find the value $\varphi^*_{\mathbb{P}}$ that optimizes $\mathbb{P}$ subject to a maximum error rate \errorratemax{}. The algorithm aims to solve the following optimization problem:
\begin{equation}
    \text{maximise}_\varphi  \{ \mathbb{P} \} \quad \text{subject to: }  E \leq E_{max} \label{eq:optimization}
\end{equation}

where $\mathbb{P}$ is the target performance: the $F_1$-Score ($F_1$), Precision ($Pr$) or Recall ($Rec$) of the targeted class (malware in this article); \errorratemax{} is the maximum tolerated error; depending on the target $\mathbb{P}$, the error rate \errorrate{} has a different formulation:
\begin{center}
\begin{varwidth}{\textwidth}
\begin{itemize}
    \item if $\mathbb{P}=F_1 \rightarrow$ \errorrate{}$\ =1-\text{Acc}=\frac{(FP+FN)}{(TP+TN+FP+FN)}$
    \item if $\mathbb{P}=Rec \rightarrow$ \errorrate{}$\ =FPR=\frac{FP}{(TN+FP)}$
    \item if $\mathbb{P}=Pr \rightarrow$ \errorrate{}$\ =FNR=\frac{FN}{(TP+FN)}$
\end{itemize}
\end{varwidth}
\end{center}

Each of these definitions of \errorrate{} is targeted to limit the error induced by the specific performance. The choice of a particular error rate for each target performance metric, $\mathbb{P}$, is grounded in the inherent characteristics and priorities of these metrics. When the target is the $F_1$-Score ($\mathbb{P}=F_1$), the error rate is defined as $1-\text{Accuracy (Acc)}$ because the $F_1$-Score seeks a detecting balance between both class (in multi-class detection, it will be the targeted class and other classes), thereby necessitating a comprehensive view of both classification errors (FP and FN). Accuracy encapsulates this balance by considering the correct classifications (TP and TN).

\begin{center}
\begin{minipage}{0.90\textwidth}
    \begin{algorithm}[H]
        \caption{Tuning $\varphi$.}
        \footnotesize
        \label{alg:varphi}
        \DontPrintSemicolon
        \SetAlgoNoEnd
        \SetKwInOut{Parameter}{Parameters}
    
        \KwIn{Training dataset $Tr$}
        
        \Parameter{Learning rate $\mu$, target performance $\mathbb{P} \in \{F_1, Pr, Rec\}$, max error rate \errorratemax{}}
        \KwOut{$\varphi^{*}_{\mathbb{P}}$, optimal percentage of mw to use in training to achieve the best target performance $\mathbb{P}$ subject to \errorrate{}$<$\errorratemax{}.}
    
        \vspace{0.25em}
    
        Split the training set \emph{Tr} into two subsets: actual training (\emph{ProperTr}) and validation set (\emph{Val}), while enforcing C1, C2, C3 (\S~\ref{subsec:constrains}), also implying $\delta=\hat{\sigma}$\label{alg:varphi:init-start}\;
    
        Divide \emph{Val} into $N$ non-overlapped subsets, each corresponding to a time-slot $\Delta$, so that \emph{Val}$_{array}$ = $[V_0, V_1,...,V_N]$ \;
    
        Train a classifier $\Theta$ on \emph{ProperTr}\;
    
        \performancestar{} $\gets$ AUT($\mathbb{P}$,$N$) on \emph{Val}$_{array}$ with $\Theta$\;
    
        $\varphi^*_{\mathbb{P}} = \hat{\sigma}$ \label{alg:varphi:init-end}\;
        \For{($\varphi=\mu$; $\varphi < 1.0$; $\varphi = \varphi + \mu$)}{
            \label{alg:varphi:core-start}
        
            Downsample gw in \emph{ProperTr} so that percentage of mw is $\varphi$\;
            
            Train the classifier $\Theta_\varphi$ on \emph{ProperTr} with $\varphi$ mw \;
            
            performance \performance{}$_\varphi$ $\gets$ AUT($\mathbb{P}$, $N$)  on \emph{Val}$_{array}$ with $\Theta_\varphi$\;
            
            error \errorrate{}$_\varphi$ $\gets$ Error rate on \emph{Val}$_{array}$ with $\Theta_\varphi$\;
        
            \If{(\performance{}$_\varphi > $ \performancestar{}) {\bf and} (\errorrate{}$_\varphi \leq$\errorratemax{})} {
                \label{alg:varphi:error}
            	\performancestar{} $\gets $ \performance$_\varphi$\;
            	$\varphi^*_{\mathbb{P}} \gets \varphi$\;
            }
        }
        
        \Return $\varphi^*_{\mathbb{P}}$\; \label{alg:varphi:core-end}
    \end{algorithm}
\end{minipage}
\end{center}

In contrast, when targeting Recall ($\mathbb{P}=Rec$), the error rate is focused on the FPR. This is because when we improve a classifier's ability to catch positive class, the side effect, which we consider as errors, will be the increase of false positives (FP). FPR directly measures the extent to which a model incorrectly classifies negatives as positives. Similarly, when Precision ($\mathbb{P}=Pr$) is the goal, the error rate is defined by the False Negative Rate ($\mathit{FNR}$) because in this scenario, the errors will be positive samples which are wrongly detected as negative.

As a result, if we want to maximize $F_1$-Score for the malware class, we need to limit both FPs and FNs. if $\mathbb{P}=Pr$, we minimize FNs, so we constrain FNR. And if $\mathbb{P}=Rec$, we constrain FPR.

Algorithm~\ref{alg:varphi} consists of two phases: \emph{initialization} (lines~\ref{alg:varphi:init-start}--\ref{alg:varphi:init-end}) and \emph{grid search} of $\varphi^*_{\mathbb{P}}$ (lines \ref{alg:varphi:core-start}--\ref{alg:varphi:core-end}). In the initialization phase, the training set $Tr$ is split into a proper training set \emph{ProperTr} and a validation set \emph{Val}; this split is according to the space-time evaluation constraints in \S~\ref{subsec:constrains}, so that all the objects in \emph{ProperTr} are temporally anterior to \emph{Val}, and the malware percentage $\delta$ in \emph{Val} is equal to $\hat{\sigma}$, the in-the-wild malware percentage. The maximum performance observed \performancestar{} and the optimal training ratio $\varphi^*_{\mathbb{P}}$ are initialized by assuming the estimated in-the-wild malware ratio $\hat{\sigma}$ for training; in Android, $\hat{\sigma} \approx 10\%$ (see \S~\ref{subsec:Malware Ratio}).

The grid-search phase iterates over different values of $\varphi$, with a learning rate $\mu$ (e.g., $\mu=0.05$), and keeps as $\varphi^*_{\mathbb{P}}$ the value leading to the best performance, subject to the error constraint. To reduce the chance of discarding high-quality points while down-sampling goodware, we prioritize the most uncertain points (e.g., points close to the decision boundary of the SVM)~\cite{Settles:AL}. This approach differs from the one in the conference version of \tesseract~\cite{pendlebury2019tesseract}, where we relax the constraint on line~\ref{alg:varphi:core-start} from ($\hat{\sigma} \leq \varphi \leq 0.5$) to a broader range of ($\mu \leq \varphi \leq 1.0$). We find that imposing a stringent boundary is unnecessary when an error rate control is already in place (see \autoref{alg:varphi}, \autoref{alg:varphi:error}). 
Finally, the grid-search explores multiple values of $\varphi$ and stores the best ones. To capture time-aware performance, we rely on AUT (\S~\ref{subsec:performance metrics}), and the error rate is computed according to the target $\mathbb{P}$ (see above). Tuning examples are in \S~\ref{subsec:reveal performance}.

\begin{figure}[t]
\centering
    \includegraphics[width=0.95\textwidth]{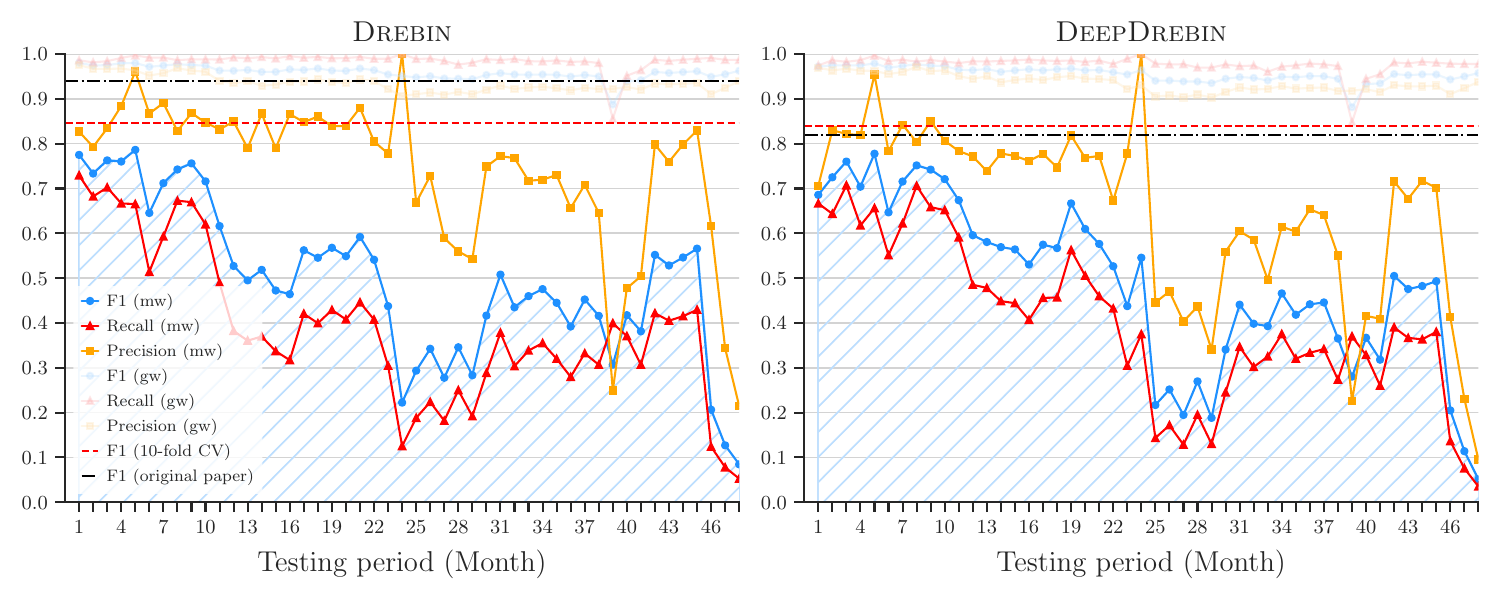}
	
	\caption{Performance decay over time of \drebin{}~\cite{Arp:Drebin} and \dl{}~\cite{Papernot:ESORICS}---with AUT$(F_1,48m)$ of 0.50 and 0.49, respectively. Training on data from 2014 and testing on data from 2015 to 2018. Both distribution have roughly 10\% malware.}
	\label{fig:drebin decay}
\end{figure}

\subsection{\tesseract: Revealing Hidden Performance}
\label{subsec:reveal performance}

Here, we show how our methodology can reveal hidden performance of \drebin~\cite{Arp:Drebin} and \dl~\cite{Papernot:ESORICS}, along with their robustness to time decay.

We introduce \tesseract, an open-source performance evaluation framework that rigorously applies constraints C1, C2, and C3 (\S~\ref{subsec:constrains}), computes AUT (\S~\ref{subsec:performance metrics}), and possesses the capability to refine a classifier utilizing our proposed tuning algorithm (\S~\ref{subsec:tuning alg}). \tesseract functions comparably to traditional Python ML libraries, yet it distinctively requires not only the features matrix~$X$ and labels~$y$, but also an array of timestamps 
~$t$, which sorts the dates corresponding to each data entry. Details on \tesseract's architecture and its versatility are thoroughly discussed in Appendix \ref{sec:artifact}.

\autoref{fig:drebin decay} reports several performance metrics of the
two selected classification approaches as point estimates over time. The $X$-axis reports 
the testing slots in month granularity, whereas the $Y$-axis reports different
scores between 0.0 and 1.0. The areas highlighted in blue correspond to
the AUT($F_1,48m$) over the 48 testing months. The black dash-dotted horizontal lines represent the best $F_1$-Score from the original
works Arp \etal \cite{Arp:Drebin} and Papernot \etal \cite{Papernot:ESORICS},
corresponding to results obtained with 10 hold-out random splits for
\drebin and \dl, respectively; both
these settings are analogous to k-fold from a temporal bias
perspective, and violate both C1 and C2. The red dashed horizontal lines correspond to 10-fold $F_1$ obtained on our dataset, which satisfies C3.

The figure illustrates that \dl's performance aligns closely with that reported in the original study by Papernot \etal \cite{Papernot:ESORICS}, as indicated by the similarity between the red-dashed and black dash-dotted lines. In contrast, \drebin exhibits a slightly reduced performance compared to its original evaluation in Arp \etal \cite{Arp:Drebin}. We believe the difference of performance is due to different data distribution. Both approaches show a similar trend of performance in the testing data. The 10-fold CV results are 0.85 and 0.84, with an AUT(\(F_1,48m\)) of 0.50 and 0.49 for \drebin and \dl, respectively. Focusing on the malware side (highlighted lines), we observe that both $F_1$-Scores and Recall rates initially decrease, followed by a modest recovery after Month 28. Precision remains stable in the first 24 months for both methods, then experiences a notable decline after Month 24 and generally remains lower thereafter. There is no obvious performance decay, apart from Month 39, from goodware's side (dimmed lines).

Notably, between these two detection approaches, \dl demonstrates greater robustness, especially in the second year of the testing period, achieving higher Recall and $F_1$-Score. We present a more detailed AUT analysis in \autoref{table:AUTCompare}.


\begin{table}[t]
    \centering
    \caption{AUT comparison across various observation windows ($\tau$) for the \drebin and \dl approaches. Rows under the 'AUT Observation Window $\tau$' show the size of selected observation windows. Each column under the 'Testing Month' cell indicates the endpoint of the testing period, starting from the end of the preceding cell. For instance, when $\tau$ is 3-month, the cell under column '12' denotes the AUT result for the window spanning months 10 to 12. Similarly, for the row of $\tau$ is 6-month, the cell under column '12' corresponds to the AUT result from months 7 to 12. Time windows where one model significantly outperforms the other, determined by a minimum AUT difference of $0.02$, are highlighted - orange for \drebin and blue for \dl. Note that this table's $\mathbb{P}$ is $F_1$-Scores, and the AUT granularity is a month.}
    \vspace{1mm}
    \begin{adjustbox}{width=\textwidth}
        \begin{tabular}{|ll||l|l|l|l|l|l|l|l|l|l|l|l|l|l|l|l|}
            \hline
            \multicolumn{18}{|c|}{\textbf{Time-Aware AUT Comparison}} \\
            \hhline{|==================|}
            \multicolumn{2}{|c||}{\multirow{2}{*}{AUT Observation Window $\tau$}} & \multicolumn{16}{c|}{Testing Month} \\
            \cline{3-18}
            \multicolumn{2}{|l||}{} & 3 & 6 & 9 & 12 & 15 & 18 & 21 & 24 & 27 & 30 & 33 & 36 & 39 & 42 & 45 & 48 \\ 
            \hhline{|==||=|=|=|=|=|=|=|=|=|=|=|=|=|=|=|=|}
            \multicolumn{1}{|l|}{\multirow{2}{*}{3-month}} & \drebin & \multicolumn{1}{>{\columncolor{orange!50}\hsize=1\hsize}c|}{0.75}& \multicolumn{1}{>{\columncolor{orange!50}\hsize=1\hsize}c|}{0.74}& \multicolumn{1}{>{\columncolor{orange!50}\hsize=1\hsize}c|}{0.74}& \multicolumn{1}{c|}{0.62}& \multicolumn{1}{c|}{0.50}& \multicolumn{1}{c|}{0.53}& \multicolumn{1}{c|}{0.57}& \multicolumn{1}{c|}{0.41}& \multicolumn{1}{>{\columncolor{orange!50}\hsize=1\hsize}c|}{0.31}& \multicolumn{1}{>{\columncolor{orange!50}\hsize=1\hsize}c|}{0.33}& \multicolumn{1}{>{\columncolor{orange!50}\hsize=1\hsize}c|}{0.46}& \multicolumn{1}{c|}{0.44}& \multicolumn{1}{>{\columncolor{orange!50}\hsize=1\hsize}c|}{0.40}& \multicolumn{1}{>{\columncolor{orange!50}\hsize=1\hsize}c|}{0.43}& \multicolumn{1}{>{\columncolor{orange!50}\hsize=1\hsize}c|}{0.55}& \multicolumn{1}{c|}{0.14} \\
            \cline{2-18} 
            \multicolumn{1}{|l|}{} & \dl & \multicolumn{1}{c|}{0.72}& \multicolumn{1}{c|}{0.70}& \multicolumn{1}{c|}{0.69}& \multicolumn{1}{>{\columncolor{blue!50}\hsize=1\hsize}c|}{0.67}& \multicolumn{1}{>{\columncolor{blue!50}\hsize=1\hsize}c|}{0.59}& \multicolumn{1}{>{\columncolor{blue!50}\hsize=1\hsize}c|}{0.56}& \multicolumn{1}{>{\columncolor{blue!50}\hsize=1\hsize}c|}{0.60}& \multicolumn{1}{c|}{0.41}& \multicolumn{1}{c|}{0.23}& \multicolumn{1}{c|}{0.26}& \multicolumn{1}{c|}{0.44}& \multicolumn{1}{c|}{0.44}& \multicolumn{1}{c|}{0.37}& \multicolumn{1}{c|}{0.41}& \multicolumn{1}{c|}{0.52}& \multicolumn{1}{c|}{0.13}\\ 
            \hhline{|==||==|==|==|==|==|==|==|==|}
            \multicolumn{1}{|l|}{\multirow{2}{*}{6-month}} & \drebin & \multicolumn{2}{>{\columncolor{orange!50}\hsize=2\hsize}c|}{0.75}& \multicolumn{2}{c|}{0.69}& \multicolumn{2}{c|}{0.51}& \multicolumn{2}{c|}{0.50}& \multicolumn{2}{>{\columncolor{orange!50}\hsize=2\hsize}c|}{0.32}& \multicolumn{2}{>{\columncolor{orange!50}\hsize=2\hsize}c|}{0.46}& \multicolumn{2}{>{\columncolor{orange!50}\hsize=2\hsize}c|}{0.40}& \multicolumn{2}{c|}{0.35} \\ 
            \cline{2-18} 
            \multicolumn{1}{|l|}{} & \dl & \multicolumn{2}{c|}{0.71}& \multicolumn{2}{c|}{0.69}& \multicolumn{2}{>{\columncolor{blue!50}\hsize=2\hsize}c|}{0.57}& \multicolumn{2}{>{\columncolor{blue!50}\hsize=2\hsize}c|}{0.52}& \multicolumn{2}{c|}{0.24}& \multicolumn{2}{c|}{0.44}& \multicolumn{2}{c|}{0.38}& \multicolumn{2}{c|}{0.33} \\ 
            \hhline{|==||====|====|====|====|}
            \multicolumn{1}{|l|}{\multirow{2}{*}{12-month}} & \drebin & \multicolumn{4}{>{\columncolor{orange!50}\hsize=4\hsize}c|}{0.72}& \multicolumn{4}{c|}{0.51}& \multicolumn{4}{>{\columncolor{orange!50}\hsize=4\hsize}c|}{0.39}& \multicolumn{4}{>{\columncolor{orange!50}\hsize=4\hsize}c|}{0.39} \\ 
            \cline{2-18} 
            \multicolumn{1}{|l|}{} & \dl & \multicolumn{4}{c|}{0.69}& \multicolumn{4}{>{\columncolor{blue!50}\hsize=4\hsize}c|}{0.55}& \multicolumn{4}{c|}{0.35}& \multicolumn{4}{c|}{0.37}\\ 
            \hhline{|==||========|========|}
            \multicolumn{1}{|l|}{\multirow{2}{*}{24-month}} & \drebin & \multicolumn{8}{c|}{0.61}& \multicolumn{8}{>{\columncolor{orange!50}\hsize=8\hsize}c|}{0.39} \\ 
            \cline{2-18} 
            \multicolumn{1}{|l|}{} & \dl & \multicolumn{8}{c|}{0.62}& \multicolumn{8}{c|}{0.36} \\ 
            \hhline{|==||================|}
            \multicolumn{1}{|l|}{\multirow{2}{*}{48-month}} & \drebin & \multicolumn{16}{c|}{0.50}\\ 
            \cline{2-18} 
            \multicolumn{1}{|l|}{} & \dl & \multicolumn{16}{c|}{0.49}\\ 
            \hhline{|==||================|}
        \end{tabular}
    \end{adjustbox}
    \label{table:AUTCompare}
\end{table}

{\bf Observation window in AUT evaluation.} The nature of the AUT metric we propose shows the overall performance of a classifier in the time-aware evaluation. As a result, it sometimes ignores performance details for a specific period. For instance, even if the AUT($F_1$, 48m) for \drebin and \dl are 0.50 and 0.49, respectively, we can still observe that at some periods (e.g., first five months and last seven months), the performance can vary. 

When evaluating performance with the AUT metric, a new feature in \tesseract framework is that users can define the observation time window ($\tau$, by default, it equals the full length of the testing period) for more detailed performance change. Different from the granularity parameter, which defines the size of test time slots $\Delta$ (\autoref{eq:c2}), $\tau$ can be considered as the length of the period over which we assess the classifier's performance. This observation window is important as it shapes the scope of performance evaluation. For example, a smaller $\tau$ might involve assessments every three months, while a larger one could span several years. Adjusting $\tau$ of AUT is essential for understanding how a classifier's effectiveness in malware detection evolves over time. Different $\tau$ can lead to distinct insights, enabling researchers to pinpoint when and how a classifier's performance starts to decay, thus guiding the development of robust systems.

As shown in \autoref{table:AUTCompare}, the AUT comparison between \drebin and \dl highlights the impact of varying the AUT observation window $\tau$. The table presents AUT values across different $\tau$, ranging from 3-month to 48-month, illustrating how performance fluctuates over time. Cells highlighted in the table indicate where one model notably outperforms the other, signaling areas for further investigation. Both approaches are initialized with training data from 2014, with the AUT metric evaluated over the four years of testing data. When we set the time window as default, where $\tau =$ 48-month, \drebin and \dl show similar performances of 0.50 and 0.49, respectively. However, by reducing $\tau$ reveals variations in performance. When $\tau =$ 24-month, we can find that both approaches perform similarly in the two evaluation periods, with a better AUT in the first 24 months. Notably, when $\tau$ gets even smaller, \dl exhibits more consistent and robust performance in the second year (Month 13 to 24) and maintains this trend across all smaller time windows (e.g., $\tau =$ 3-month, 6-month, and 12-month).

This variability in smaller windows suggests that $\tau$ can provide deeper insights into a classifier's immediate adaptability and response to emerging malware types. Larger windows might offer a broader view of long-term stability and resilience. Shorter windows are beneficial in environments where malware threats evolve rapidly, and early detection is crucial. These findings highlight the significance of selecting the appropriate AUT observation window $\tau$. 

The table also highlights another important aspect: the need to investigate the causes behind the performance discrepancies at different granularities. Such an investigation could involve Explainable AI (XAI) techniques, which can provide insights into the decision-making processes of classifiers \cite{Chow:DriftForensics}. Understanding why a classifier's performance varies with time can inform improvements in classifier design and contribute to developing more robust malware detection systems.

In practice, the selection of $\tau$ should be driven by the specific needs of the malware detection context and the desired balance between immediate responsiveness and enduring robustness. The capability to compute AUT over different windows demonstrates its versatility and enables a more strategic and customized approach to classifier evaluation and comparison.

\textbf{Violating the consistency of temporal training (C1) and gw/mw time-window (C2).} The removal of temporal bias unveils the real performance of each algorithm amid concept drift. The AUT(\(F_1\), 48m) captures such performance over all 48 testing months: 0.50 for \drebin, and 0.49 for \dl, respectively. We also evaluate the two approaches on the first 24 months of testing data: the AUT(\(F_1\), 24m) stands at 0.61 for \drebin and 0.64 for \dl, respectively. Across all quartets of scenarios, the AUT consistently falls behind the 10-fold \(F_1\) score, as the latter violates constraint C1 and may violate C2 if the dataset classes are not evenly distributed across the timeline.

\textbf{Robustness of deep learning approach.} In our initial analysis documented in the original paper (refer to the `DL' plot of Figure 5 in \cite{pendlebury2019tesseract}), the deep learning method appeared to outshine the other two traditional ML approaches (\texttt{Alg1(\drebin)} and \texttt{Alg2(\mmd)}) in terms of robustness after enforcing C1, C2 and C3. However, upon closer inspection and a more extended evaluation, this claim does not hold consistently. It is true that during the first two year of evaluation, the deep learning approach demonstrated a better and relatively stable performance. Nonetheless, when considering the broader scope of the analysis, it becomes evident that the overall performance of the deep learning method aligns closely with that of traditional machine learning techniques, if retraining is not possible. This revelation underscores that while deep learning may offer advantages in certain temporal slices, its efficacy is comparable to traditional models when viewed across a comprehensive timeline. This unresolved inquiry is delegated to future investigative efforts to discern the conditions under which \dl, or deep learning in general, exhibits superior performance within certain testing intervals.

\begin{table}[t]
    \centering
    
    \caption{Testing AUTs performance when training with $\hat{\sigma}$, $\varphi^*_{F_1}$, $\varphi^*_{Pr}$, and $\varphi^*_{Rec}$. Results are shown for both the entire 48-month period and the first 24-month period, offering a comprehensive view of performance over different time frames. Best performances are highlighted in gray color.}

    \vspace{1mm}
    
    \resizebox{\columnwidth}{!}{%
      
        \begin{tabular}{|l|l|r|r|c|c|c||r|r|c|c|c|}
        \hline
        \multirow{2}{*}{{\bf Algorithm}}&
        \multirow{2}{*}{\textbf{$\varphi$}} &
        \multirow{2}{*}{FP (48m)} &
        \multirow{2}{*}{FN (48m)} &
        \multicolumn{3}{c|}{\textbf{AUT($\mathbb{P}$, 48m)}} &
        \multirow{2}{*}{FP (24m)} &
        \multirow{2}{*}{FN (24m)} &
        \multicolumn{3}{c|}{\textbf{AUT($\mathbb{P}$, 24m)}} \\
        \hhline{~~~~---~~---}
    
        &
        &  &  & $F_1$ & $Pr$ & $Rec$ 
        &  &  & $F_1$ & $Pr$ & $Rec$ \\
        \hline \hline

        \multirow{4}{*}{\drebin{}~\cite{Arp:Drebin}}
            & $\hat{\sigma}$ ($\sim$10\%)       & 3,201  & 13,251 & 0.50 & \cellcolor{lightgray!90} 0.75 & 0.39 & 596 & 3,716 & 0.61 & 0.85 & 0.49 \\
            & $\varphi^{*}_{F_1}$ (55\%)
            & 12,053 & 10,409 & \cellcolor{lightgray!90} 0.51 & 0.49 & 0.54  & 2,970 & 2,113 & \cellcolor{lightgray!90} 0.64 & 0.61 & 0.69 \\
            & $\varphi^{*}_{Pr}$ (5\%)       & 2,616 & 15,637 & 0.39 &  0.72 & 0.28 & 421 & 4,025 & 0.58 & \cellcolor{lightgray!90}0.87 & 0.45 \\
            & $\varphi^{*}_{Rec}$ (95\%)     & 48,891 & 5,878 & 0.37 & 0.25 & \cellcolor{lightgray!90} 0.75 & 13,978 & 466 & 0.46 & 0.31 & \cellcolor{lightgray!90} 0.91 \\
            \hline \hline
    
        \multirow{4}{*}{\dl{}~\cite{Papernot:ESORICS}}
            & $\hat{\sigma}$ ($\sim$10\%)       & 4,711 & 13,180 & 0.49 & 0.65 & 0.40 & 989 & 3,292 & 0.64 & 0.80 & 0.54 \\
            & $\varphi^{*}_{F_1}$ (20\%)
            & 6,381  & 11,509 & \cellcolor{lightgray!90} 0.53 & 0.62 & 0.48 & 1,494 & 2,605 & \cellcolor{lightgray!90}0.68 & 0.74 & 0.63 \\
            & $\varphi^{*}_{Pr}$ (5\%)      & 2,069 & 16,961 & 0.32 & \cellcolor{lightgray!90} 0.69 & 0.22 & 302 & 4,792 & 0.47 & \cellcolor{lightgray!90}0.89 & 0.33 \\
            & $\varphi^{*}_{Rec}$ (95\%)     & 86,487 & 1,844 & 0.30 & 0.18 & \cellcolor{lightgray!90} 0.91 & 28,364 & 165 & 0.32 & 0.19 & \cellcolor{lightgray!90} 0.95 \\
        
        \hline
    
        \end{tabular}
    }
      
\label{tab:varphi}
\end{table}

\begin{figure}[t]
	\centering
	\begin{subfigure}{0.32\textwidth}
		\includegraphics[width=\linewidth]{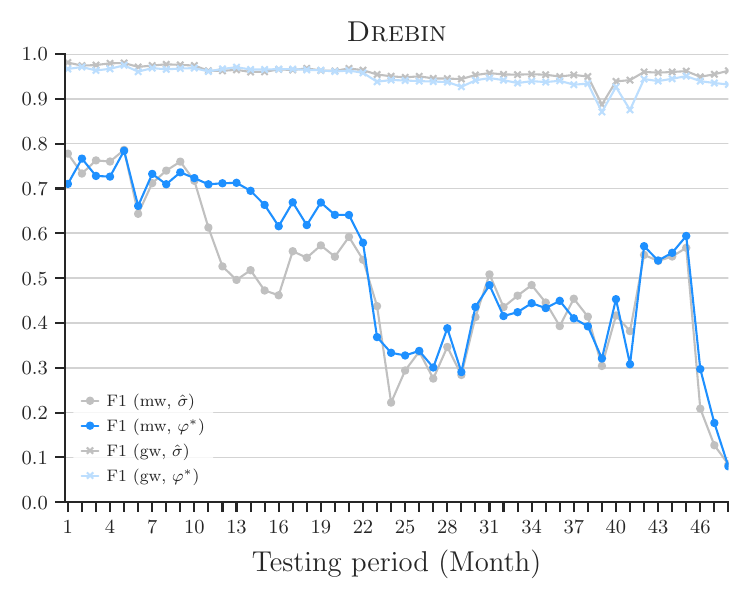}
		\caption{Tr: 2014, \( \varphi = 0.55 \)  Ts: 2015-2018 }
        \label{fig:drebin full varphi}
	\end{subfigure}
    \hfill
	\begin{subfigure}{0.32\textwidth}
		\includegraphics[width=\linewidth]{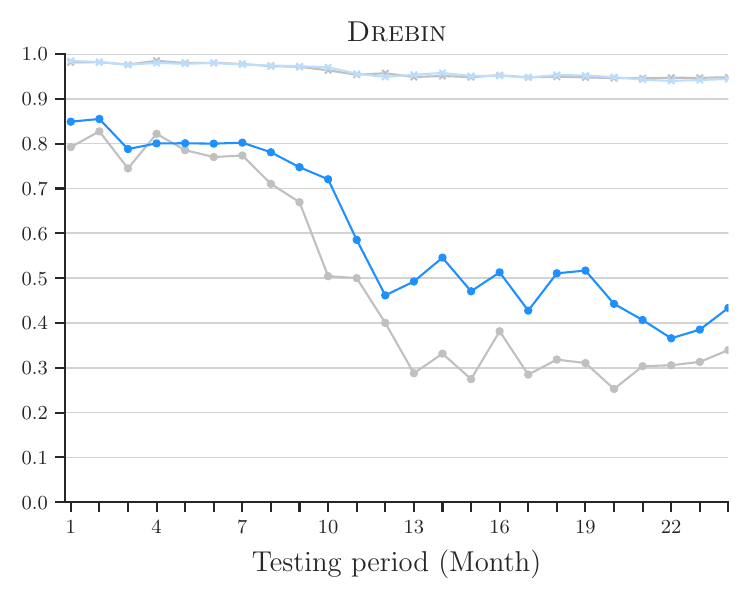}
		\caption{Tr: 2015, \( \varphi = 0.35 \)  Ts: 2016-2017}
        \label{fig:drebin 2015 varphi}
	\end{subfigure}
	\hfill
	\begin{subfigure}{0.32\textwidth}
		\includegraphics[width=\linewidth]{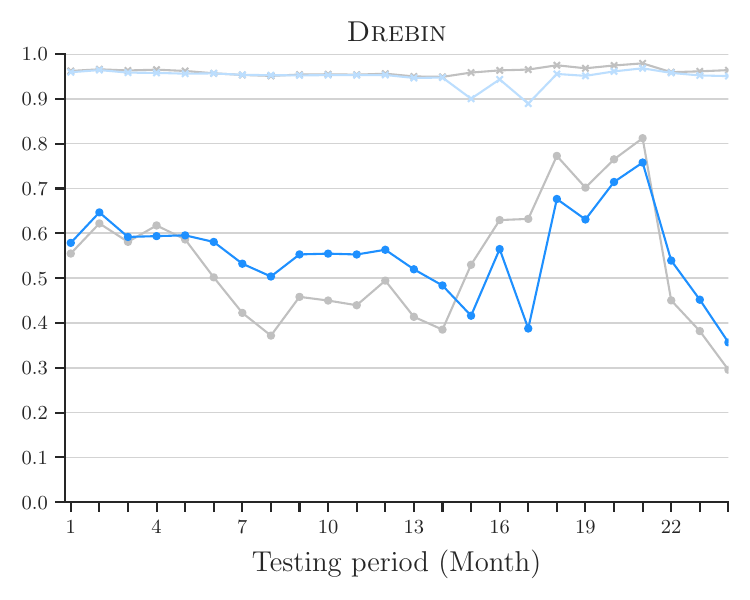}
		\caption{Tr: 2016, \( \varphi = 0.35 \)  Ts: 2017-2018}
        \label{fig:drebin 2016 varphi}
	\end{subfigure}

    \vspace{1em} 

	\begin{subfigure}{0.32\textwidth}
		\includegraphics[width=\linewidth]{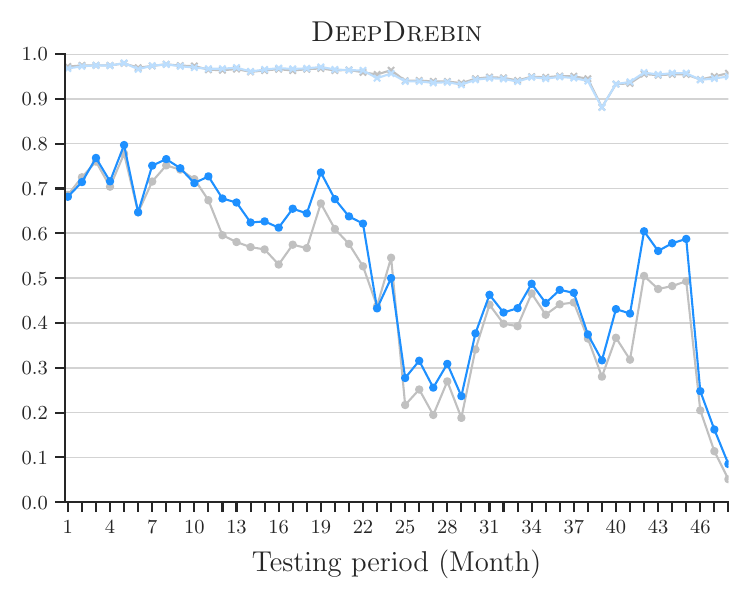}
		\caption{Tr: 2014, \( \varphi = 0.2 \)  Ts: 2015-2018}
        \label{fig:dl full varphi}
	\end{subfigure}
	\hfill
	\begin{subfigure}{0.32\textwidth}
		\includegraphics[width=\linewidth]{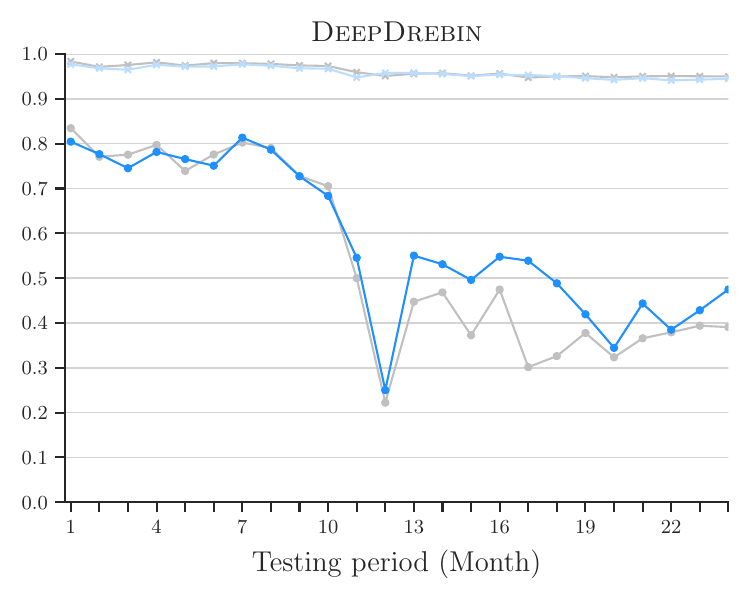}
		\caption{Tr: 2015, \( \varphi = 0.35 \)  Ts: 2016-2017}
        \label{fig:dl 2015 varphi}
	\end{subfigure}
	\hfill
	\begin{subfigure}{0.32\textwidth}
		\includegraphics[width=\linewidth]{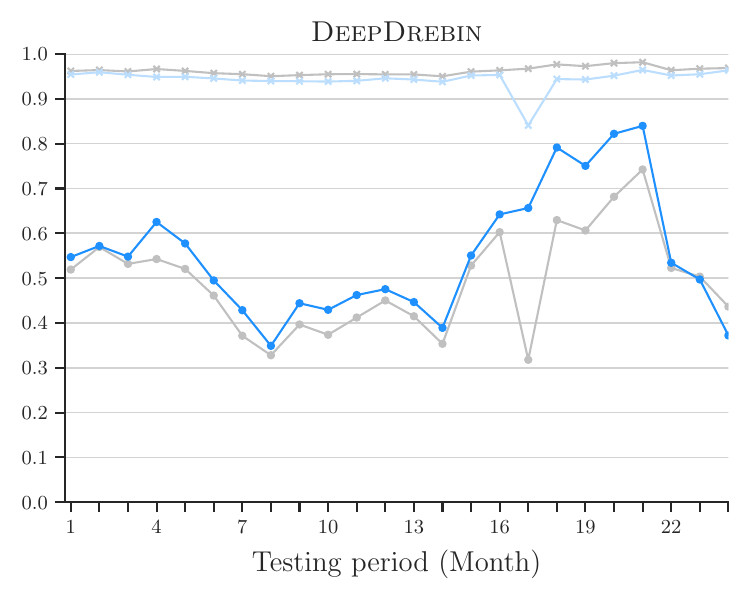}
		\caption{Tr: 2016, \( \varphi = 0.35 \)  Ts: 2017-2018}
        \label{fig:dl 2016 varphi}
	\end{subfigure}

	\caption{Performance comparison of \drebin and \dl before and after applying Tuning Algorithm (\autoref{alg:varphi}). Each graph displays results from both malware (mw) and goodware (gw) perspectives. The gray lines indicate the original $F_1$-Score performance, while the blue lines show the $F_1$-Score after tuning. The tuned outcomes in each scenario demonstrate improved performance over the original ones, achieving either better performance or increased stability. The sub-captions for each graph detail the period of training (Tr) and testing (Ts) data used. It is important to note that the final four months of data in the training set are utilized as the validation set for the tuning algorithm. Additionally, each sub-caption specifies the optimal malware percentage $\varphi = \varphi^{*}_{F_1}$ identified during the training phase.}
 
	\label{fig:varphi}
\end{figure}

\textbf{Tuning algorithm.} We now evaluate whether our tuning algorithm (Algorithm~\ref{alg:varphi} in \S~\ref{subsec:tuning alg}) improves robustness to time decay of a malware classifier for a given target performance. We first aim to maximize $\mathbb{P}=F_1$-Score of malware class, subject to \errorratemax{}$=10\%$. After running Algorithm~\ref{alg:varphi} on \drebin{}~\cite{Arp:Drebin} and~\dl{}~\cite{Papernot:ESORICS}, we set $\varphi^{*}_{F_1}=0.55$ for \drebin{}, and $\varphi^{*}_{F_1}=0.20$ for \dl{}. Figures~\ref{fig:drebin full varphi} and~\ref{fig:dl full varphi} report the respective improvements in test performance of the two approaches on four years of testing data $Ts$ after applying $\varphi^{*}_{F_1}$ to the full training set $Tr$. 

We remark that the choice of $\varphi^{*}_{F_1}$ uses only training information (see Algorithm~\ref{alg:varphi}) and no test information is used---the optimal value is chosen from a 4-month validation set extracted from the one year of training data; this is to simulate a realistic deployment setting in which we have no a priori information about testing. \autoref{fig:drebin full varphi} (\drebin) and \autoref{fig:dl full varphi} (\dl) show that our approach for finding the best $\varphi^{*}_{F_1}$ improves the $F_1$-Score on malware at most of the four-year testing time, at the cost of slightly reduced goodware performance. \autoref{tab:varphi} shows details of how total FPs, total FNs, and AUT changed by training \drebin{} and \dl{} with $\varphi^{*}_{F_1}$, $\varphi^{*}_{Prec}$, and $\varphi^{*}_{Rec}$ instead of $\hat{\sigma}$. These training ratios have been computed subject to $E_{max} = 5\%$ for $\varphi^{*}_{Rec}$, $E_{max} = 10\%$ for $\varphi^{*}_{F_1}$, and $E_{max} = 15\%$ for $\varphi^{*}_{Prec}$; the difference in the maximum tolerated errors is motivated by the class imbalance in the dataset---which causes lower FPR and higher FNR values (see definitions in \S~\ref{subsec:tuning alg}), as there are many more goodware than malware.

\autoref{tab:varphi} details the performance over the 48-month period when employing various training strategies for the \drebin and \dl algorithms. When trained with $\varphi^{*}_{F_1}$, a decrease in Precision is noted due to an increase in False Positives (FPs), while an increase in Recall is observed as a result of a decrease in False Negatives (FNs). This trade-off leads to a slight improvement of AUT for both classifiers. For each specific performance metric $\mathbb{P}$, the tailored performance AUT($\mathbb{P}$, 48m) with $\varphi^{*}_{\mathbb{P}}$ generally surpasses the performance achieved using the original training ratio $\hat{\sigma}$ for both \drebin and \dl. The notable exception is when focusing on Precision; in this case, the tuning slightly underperforms compared to the original training ratio. This discrepancy might be attributed to the concept shift over the long-term evaluation.

To provide a more detailed perspective, AUT($\mathbb{P}$, 24m) metrics are also presented, which consistently show that the results of tuning are superior to those obtained when training with 
$\hat{\sigma}$. It is also noteworthy that the AUT for Precision may vary slightly even with a similar number of total FPs; this is due to AUT($Pr$)'s sensitivity to the timing of the occurrence of FPs. A corresponding observation holds true for the total FNs and AUT for Recall. Post-tuning, \dl's performance in terms of the F1-Score, as measured by AUT, remains comparatively higher, with a AUT($F_1$, 24m) of 0.68 compares to \drebin's 0.64. This demonstrates the effectiveness of the tuning approach, especially over the first 24-month period of evaluation, where the improvements over training with $\hat{\sigma}$ are more pronounced.

Furthermore, a three-year evaluation window was also employed to assess the tuning algorithm, as evidenced in Figures~\ref{fig:drebin 2015 varphi}, \ref{fig:drebin 2016 varphi} for \drebin{}, and Figures~\ref{fig:dl 2015 varphi}, \ref{fig:dl 2016 varphi} for \dl{}. This shorter period of evaluation suggests the necessity for regular updates to the tuning process to keep pace with the dynamic nature of malware threats. Aligning with the full dataset protocol, the first year of data from this three-year period was used to ascertain the optimal $\varphi^{*}_{F_1}$, and subsequent tests were conducted over the following two years (as listed in each sub-captions). The optimal $\varphi^{*}_{F_1}$ was found to be 0.35 on both years 2015 and 2016 for both \drebin{} and \dl{}. Compared to the baseline performance (indicated by gray lines in the graphs), the tuned outcomes demonstrate enhanced robustness in the latter two years, comparing to in the long-run evaluation.

These results highlight the effectiveness of employing a short-term evaluation interval for both tuning and testing our algorithm. The graphical analyses, as demonstrated in Figures~\ref{fig:drebin 2015 varphi}, \ref{fig:drebin 2016 varphi}, \ref{fig:dl 2015 varphi}, and \ref{fig:dl 2016 varphi}, reveal that our tuning algorithm not only enhances the overall robustness of the malware classifiers (\drebin and \dl) but also contributes to more consistent performance over time. This is particularly evident in the post-tuning improvements in the $F_1$-Score, Precision, and Recall metrics across different time frames. The tuning process, guided by Algorithm~\ref{alg:varphi}, effectively addresses the challenge of time decay, as shown by the improved AUT metrics over various periods.

Moreover, the findings highlight the importance of regular re-calibration of the tuning parameters to adapt to the evolving nature of malware threats. The optimal values of $\varphi^{*}_{F_1}$ obtained from the three-year data suggest that periodic adjustments in the tuning process can lead to better detection rates and more reliable classification. Such an approach is crucial in maintaining the efficacy of malware detection systems in a rapidly changing digital landscape.

\textbf{Conclusion.} Our investigation strongly recommends the adoption of a dynamic tuning and testing regimen. Researchers should consider the specific characteristics of their datasets and the prevailing trends in malware evolution when determining the frequency of tuning intervals. Regular updates and adjustments to the tuning parameters, as dictated by the nature of the threat environment and the performance goals, are key to sustaining the reliability and accuracy of malware classifiers over time.

\section{Windows and PDF Malware}
\label{sec:otherdomains}

We now discuss the impact of the spatial and temporal bias on ML-driven detection performance in other domains: Windows Portable Executables (PEs) and PDFs. Before diving into the results, we provide reflection and intuitions on the relationship between Android and these other domains. 

\textbf{Comparing Android and Windows PE / PDF domains.} In the realm of cybersecurity, the distinction between the Android domain and the Windows PE and PDF domains is marked and significant, especially when considering spatial and temporal aspects. The Android ecosystem, predominantly comprising mobile applications, presents a unique challenge due to its vast, diverse app market and the rapid evolution of mobile technologies. This results in a highly dynamic temporal landscape where malware characteristics and trends can shift swiftly~\cite{zhou2012dissecting, Google:Report, steglich2019revisiting}. The diversity in the source and nature of applications, ranging from official app stores to third-party platforms, adds another layer of complexity \cite{Felt2011inthewild}.
In contrast, the Windows PE and PDF domains, typically associated with desktop environments, display distinctive traits. Windows PE files, crucial for executables and libraries, are integrally linked with Windows' architecture, leading to a steadier malware evolution, in sync with the relatively longer operation system (OS) development cycle~\cite{ling2023adversarial,divya2016comparative}. PDF files, serving mainly for document exchange, exhibit a stable structure in format and usage \cite{endignoux2016caradoc, vsrndic2016hidost, gu2022novel}. Malicious activities via PDFs often exploit software and OS vulnerabilities. Intuitively, we could expect a slower onset of concept drift—along with a more gradual decline in model performance—in these domains.

\textbf{Malware ratio in the wild.} The Android ecosystem predominantly features goodware \cite{Google:Report}. Conversely, the ratio of malware to goodware in Windows PE and PDF domains is not easily quantified due to a lack of comprehensive evidence. Hence, we refer to the ratios in two popular, existing datasets: one on the \ember dataset for Windows PE files \cite{Endgame:Ember} (released by the EndGame company), and the other on the \hidost dataset for PDF files \cite{vsrndic2016hidost}.

\begin{figure}[t]
    \centering
    \begin{subfigure}{0.48\textwidth}
        \includegraphics[width=\linewidth]{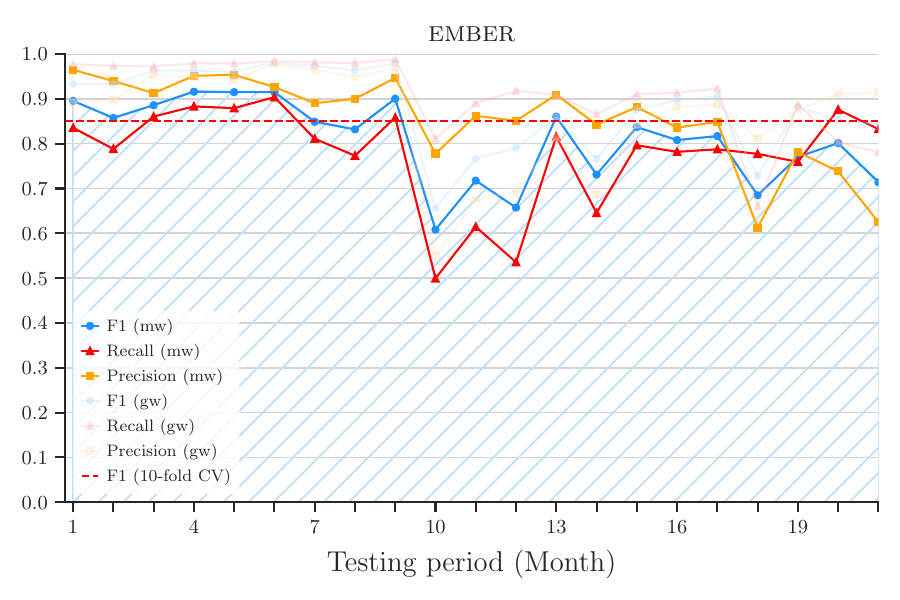}
        \vspace{-1em}
        \caption{Tr: Jan.-Mar.2017, Ts: Apr.2017 - Dec.2018.}
        \label{fig:other_domain_ember}
    \end{subfigure}
    \hfill
    \begin{subfigure}{0.48\textwidth}
        \includegraphics[width=\linewidth]{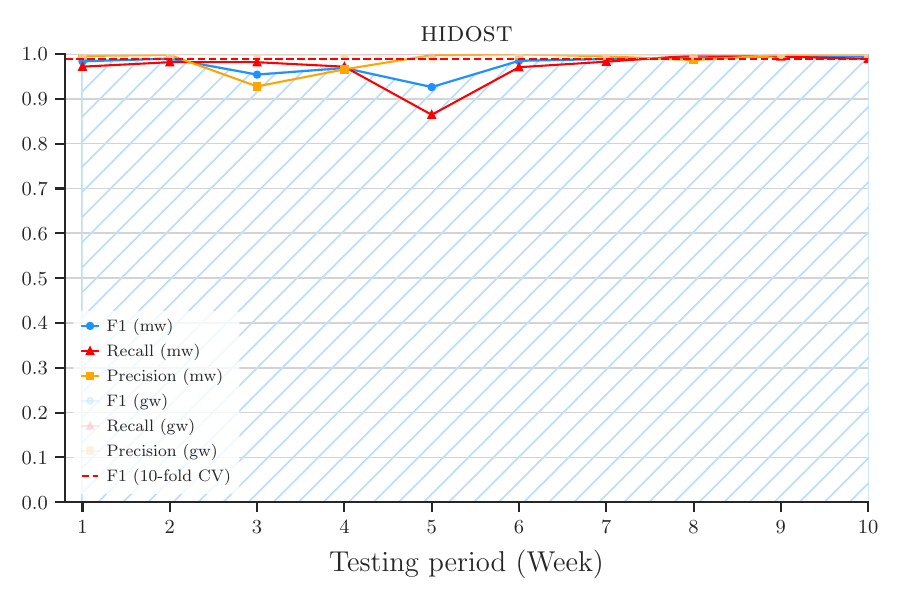}
        \vspace{-1em}
        \caption{Tr: Week 1-4, Ts: Week 5-14.}
        \label{fig:other_hidost}
    \end{subfigure}
    \caption{Time-aware evaluation on Windows PE dataset (\ember \cite{Endgame:Ember}), and PDF dataset (\hidost \cite{vsrndic2016hidost}). The Tr and Ts periods for each experiment are shown in the sub-caption.}
    \label{fig:other_domains}
\end{figure}

\begin{figure}[t]
    \centering
    \begin{subfigure}{0.48\textwidth}
        \includegraphics[width=\linewidth]{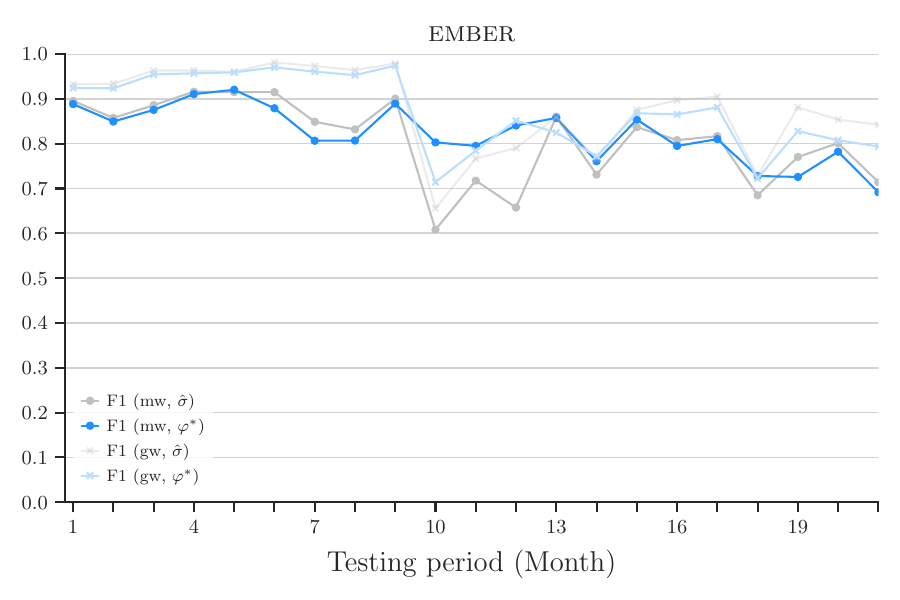}
        \vspace{-1em}
        \caption{Train on Jan 2017 and tune on the remaining training dataset, $\varphi$ = 0.25, $\varphi^*_{F_1}$ = 0.5.}
        \label{fig:other_domain_ember_tune}
    \end{subfigure}
    \hfill
    \begin{subfigure}{0.48\textwidth}
        \includegraphics[width=\linewidth]{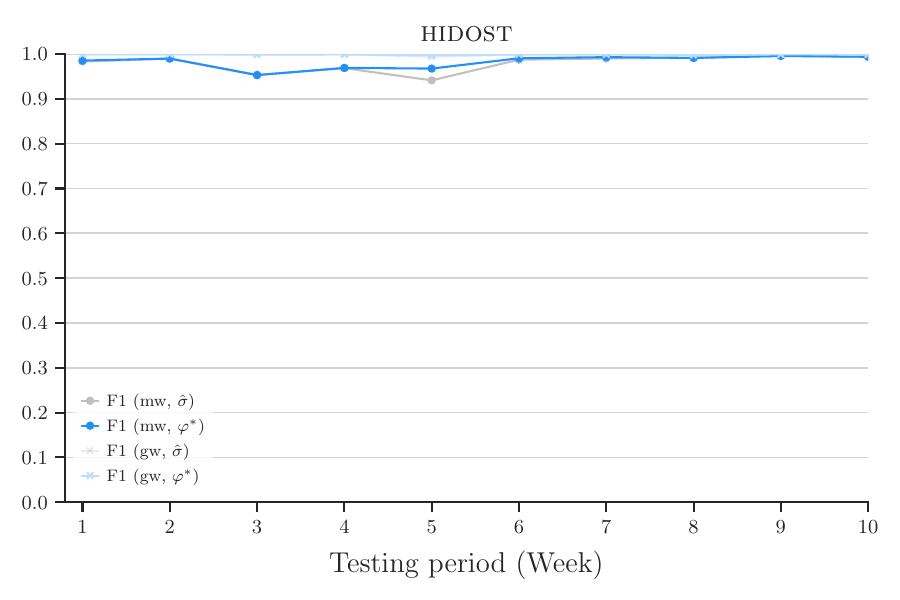}
        \vspace{-1em}
        \caption{Train on the first two weeks of the training set and tune on the remaining two. $\varphi$ = 0.04, $\varphi^*_{F_1}$ = 0.1.}
        \label{fig:other_hidost_tune}
    \end{subfigure}
    \caption{Performance comparison of \ember and \hidost before and after applying Tuning Algorithm}
    \label{fig:other_domains_tune}
\end{figure}

\subsection{Case Study: \ember}

The \ember dataset consists of a comprehensive suite of features derived from PE files to facilitate cybersecurity research \cite{Endgame:Ember}, such as headers, content-based histograms, and textual data like URL frequencies. Our analysis spans both the \ember 2017 v2 and \ember 2018 datasets, encompassing the years 2017 and 2018, resulting in a composite dataset of 255,499 benign and 148,231 malicious executables (labeled as having 40+ VirusTotal AV detections), constrained to this two-year period. We employ Gradient Boosted Decision Trees (GBDT) \cite{friedman2001greedy} as our classifier, in line with the \ember study's methods. We initialize the classifier on data from the first three months and test it on subsequent data.

The original \ember evaluation adheres to the constraints C1 and C2 delineated in \S~\ref{subsec:constrains}. Defining a precise malware-to-goodware ratio for the Windows PE domain is challenging; however, the dataset's size and heterogeneity of sources lend credence to its adherence to constraint C3. The performance degradation illustrated in~\autoref{fig:other_domain_ember} corroborates the presence of concept drift, particularly notable in the F1-score decline post-January 2018, which stabilizes at a lower benchmark. A marked reduction in the Recall metric at the onset of 2018 further underscores this point. These observations are consistent with those in the original \ember paper \cite{Endgame:Ember}, which deliberately incorporated more elusive malware into the 2018 dataset.

Applying our Tuning Algorithm on the training set improves the model's performance on the testing set, with the overall AUT($F_1$, 21m) increasing from 0.80 to 0.83. We choose the first month data from training set to initialize the tuning model, and validate the better $\varphi^*_{F_1}$ on the remaining two months data. We find the $\varphi^*_{F_1} = 0.5$, compares to the original $\varphi = 0.25$ on the training data.
Post-tuning, as shown in \autoref{fig:other_domain_ember_tune}, the model displays a consistent and robust performance, maintaining a relatively stable F1-score throughout 2018. The stability was achieved by prioritizing an increase in Recall at the expense of Precision. This approach underscores the significance of establishing a realistic malware-to-goodware ratio in the training set, which is crucial for enhancing the model's practical applicability in real-world scenarios (C3).

\subsection{Case Study: \hidost}

\hidost, an open-source tool for extracting document structures from PDF and SWF files, forms the basis of our analysis \cite{vsrndic2016hidost}. We utilize the \hidost evaluation dataset, which includes data from a 14-week period between July 16 and October 21, 2012, comprising 1,465,737 benign and 102,234 malicious files (identified with 5+ VirusTotal AV detections). The dataset's features are derived from static analysis of the PDF files, extracting structural paths in their hierarchy, resulting in both Boolean and numeric features. Following the original \hidost study, we employ a Random Forest (RF) algorithm \cite{breiman2001random}, training it on data from the initial four weeks and testing on the subsequent ten weeks.

The \hidost evaluation adheres to constraints C1 and C2. Defining a realistic malware-to-goodware ratio in the PDF domain, akin to the Windows PE domain, remains challenging. Therefore, we rely on the class balance in the original dataset. Figure \autoref{fig:other_hidost} displays stable performance across ten testing weeks, with minor dips in weeks 3 and 5. Two primary factors could explain the stable performance observed. Firstly, the robustness of the feature space is a significant contributor, especially considering that many malicious attacks in PDFs are executed via embedded JavaScript code, making them potentially more identifiable. Secondly, the limited duration of the evaluation period might not be sufficient for a concept drift to manifest significantly. This hypothesis necessitates further in-depth research to understand the true extent and nature of any potential concept drift in this context.

For tuning on the \hidost dataset, we use \autoref{alg:varphi} by initializing the model with the first two weeks of the training set, and determining the optimal $\varphi^*_{F_1} = 0.1$ on the subsequent two weeks (compared to the original $\varphi = 0.04$). Figure \autoref{fig:other_hidost_tune} illustrates the post-tuning changes. While the overall improvement is modest due to the already high baseline performance, the tuning eradicates the performance drop in week 5 and slightly improves the overall AUT($F_1$, 10w) from 0.97 to 0.98, underscoring the benefits of adjusting the malware-to-goodware ratio.

\subsection{Summary of Findings}

In exploring the Windows PE and PDF malware domains, we encounter the challenge of accurately defining an in-the-wild malware-to-goodware ratio, critical for training effective detection models. Our case studies on both \ember and \hidost reveal significantly less concept drift compared to the Android dataset, likely due to the more stable and predictable nature of malware evolution in these environments, as discussed in~\cite{chen2023overkill}. This stability is likely linked to the slower OS development cycle in Windows PE and the consistent format and usage in PDFs. Additionally, the criteria used for labeling an instance as malware vary across different approaches and domains. For instance, in the \ember dataset, an instance is labeled as malware if it triggers 40 or more VirusTotal alerts. Notably, we find that applying our \autoref{alg:varphi} to tune the malware-to-goodware ratio during the training process enhances detection performance in testing within these domains. This also indicates the importance of further research to define this ratio accurately and to understand the mechanisms behind the benefits of tuning.
\section{Delaying Time Decay}
\label{sec:delay}

\begin{figure}[t]
    \centering
    \includegraphics[width=\textwidth]{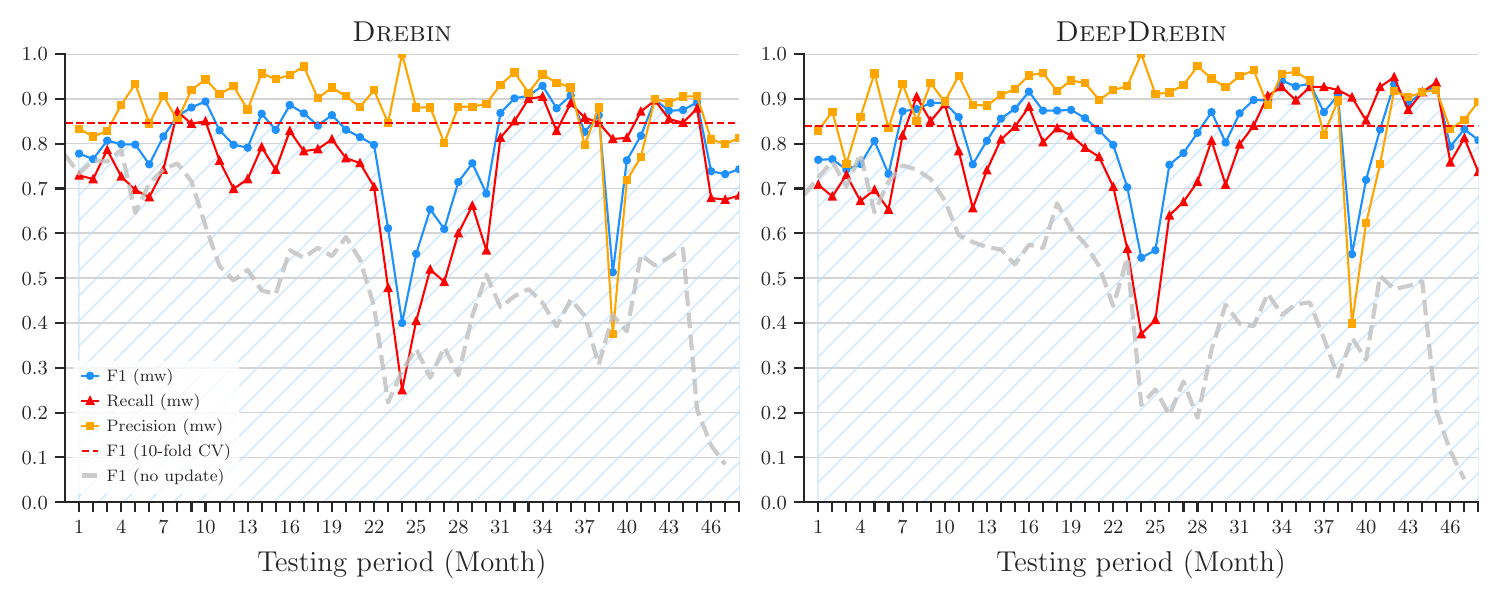}
    \caption{Delaying time decay: incremental retraining.}
    \label{fig:delay_inc_retrain}
\end{figure}

\begin{figure}[t]
    \centering
    \includegraphics[width=0.32\textwidth]{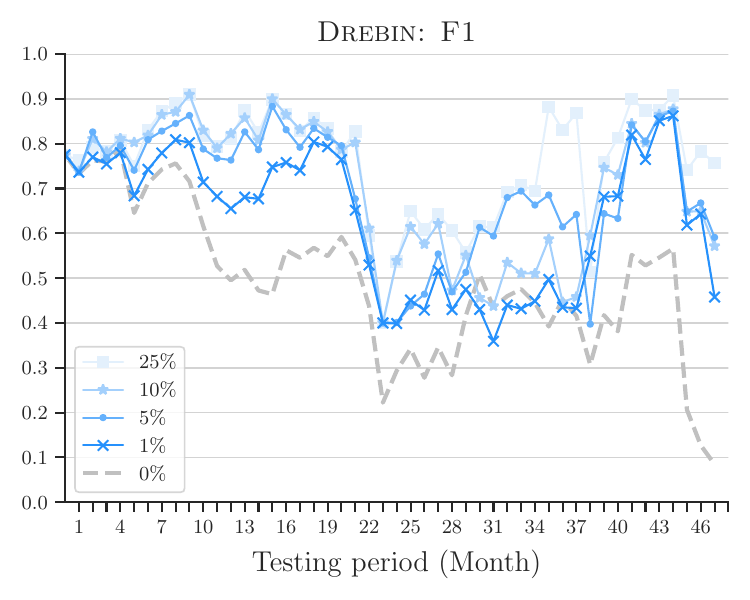}
    \includegraphics[width=0.32\textwidth]{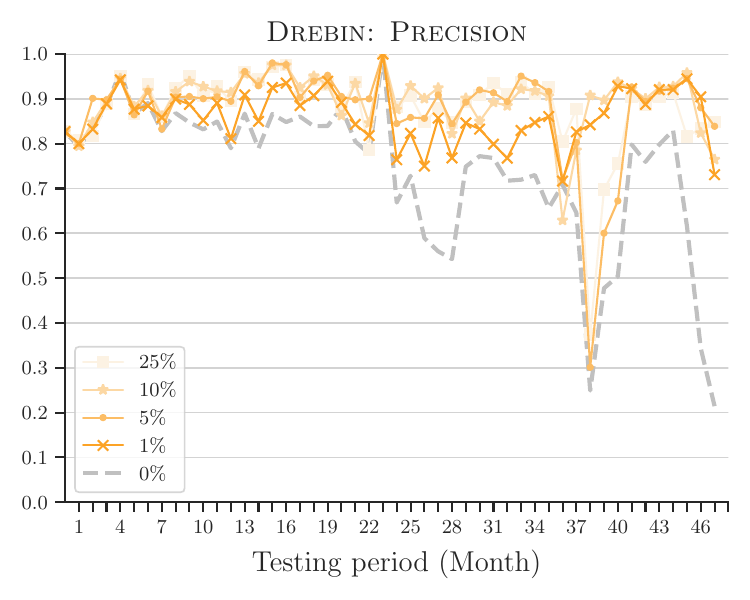}
    \includegraphics[width=0.32\textwidth]{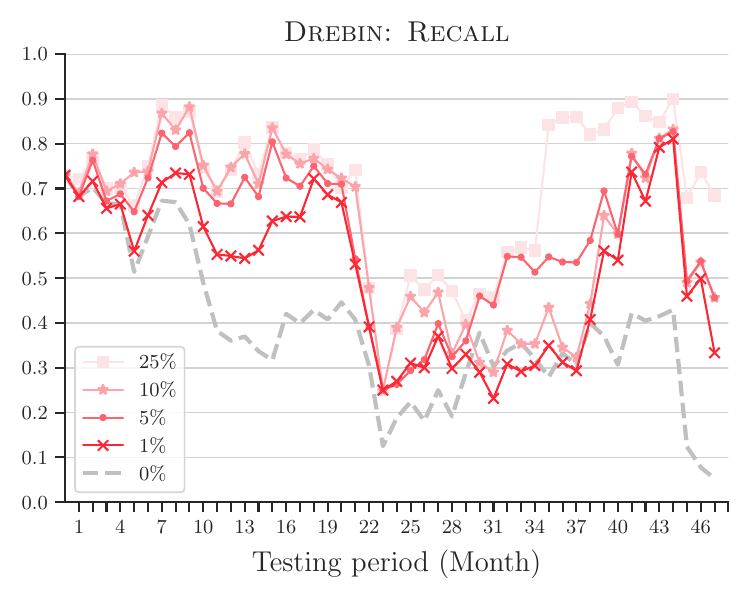}
    \newline
    \includegraphics[width=0.32\textwidth]{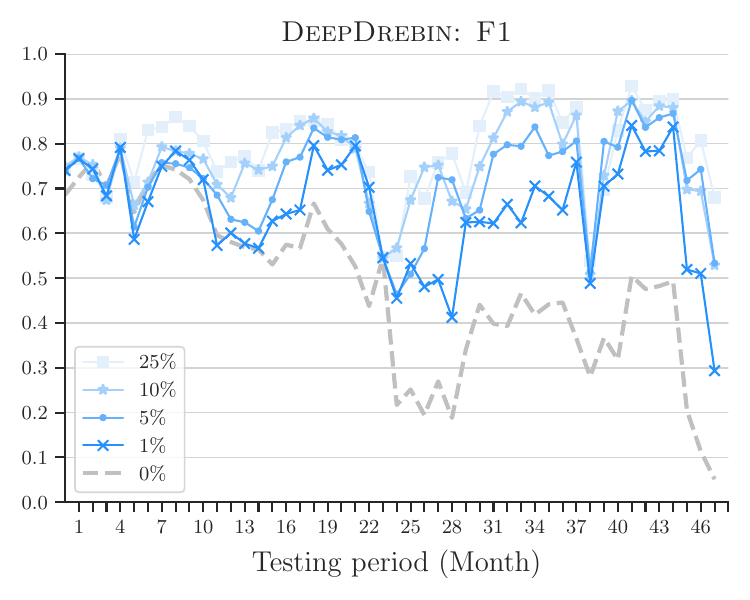}
    \includegraphics[width=0.32\textwidth]{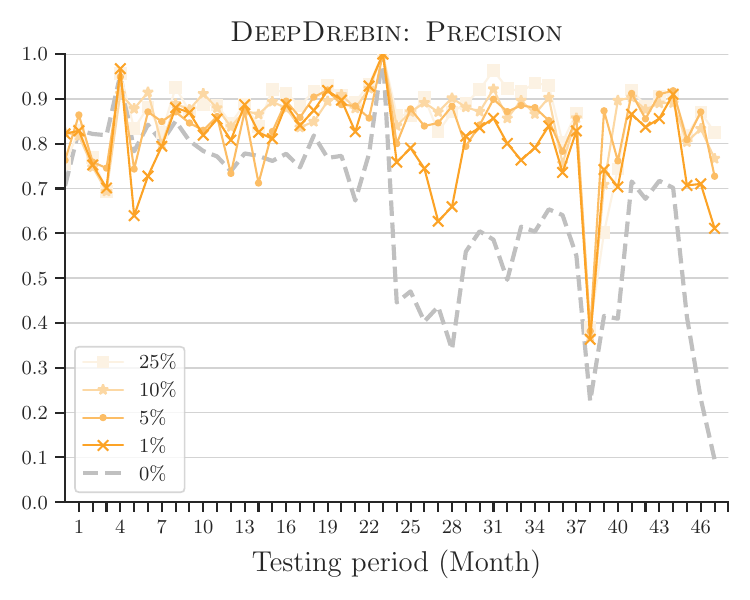}
    \includegraphics[width=0.32\textwidth]{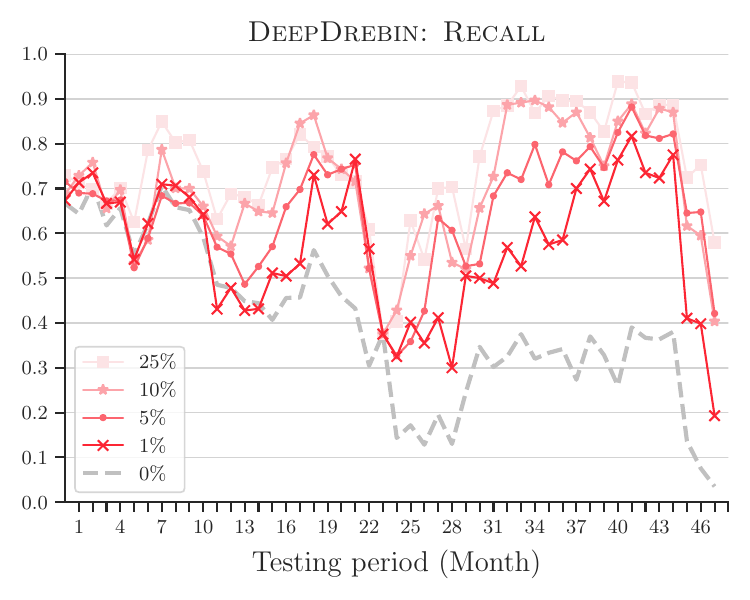}
    \caption{Delay time decay: performance with active learning based on uncertainty sampling.}
    \label{fig:delay_active_learning}
\end{figure}

In this section, we explore delay time decay strategies using the Android dataset (\S~\ref{subsec:Dataset}). We exclude the Windows PE and PDF malware datasets (\S~\ref{sec:otherdomains}) from this analysis because they suffer less from concept drift. This focus on the Android dataset, with its more pronounced data changes, provides a clearer perspective on the effectiveness of these strategies in dynamic environments.

\subsection{Delay Strategies}

While our work does not propose novel delay strategies, it leverages \tesseract to facilitate the comparative analysis of several popular approaches aimed at mitigating time decay. This provides a framework for researchers to employ \tesseract in assessing various strategies fairly, especially when proposing new solutions to delay time decay within certain budget constraints. Below, we summarize the delay strategies under consideration and present results from our dataset. Additional information on these strategies is available for interested readers.

{\bf Incremental Retraining.} We begin by examining an approach that sets an ideal performance benchmark: including \emph{all} points in a monthly retraining regime. \autoref{fig:delay_inc_retrain} illustrates the performance of \drebin and \dl with this monthly incremental retraining strategy. The results indicate a comparable upper performance limit for both methods, exhibiting an AUT($F_1$, 48m) of 0.79 for \drebin and 0.82 for \dl.  Though representing an upper bound in performance, this strategy is likely impractical due to the continuous need for labeling a vast amount of objects. Despite the potential use of VirusTotal, the increased API usage incurs additional costs. Moreover, this approach might not be suitable in various security contexts.

{\bf Active Learning.} Active Learning (AL) strategies focus on selecting a subset of test objects, whose labels are unknown, for manual labeling and inclusion in the training set. This is based on the premise that these objects are most beneficial for updating the classification model~\cite{Settles:AL}. In our study, we employ the widely-used AL query strategy known as \emph{uncertainty sampling} \cite{lewis1994heterogeneous}. Here, objects with the most uncertain predictions are chosen for retraining, under the assumption that they are key to refining decision boundaries. \autoref{fig:delay_active_learning} displays the active learning outcomes via uncertainty sampling, detailing the $F_1$ score (blue), Precision (yellow), and Recall (red) metrics. Notably, the data indicates discernible performance gains with the retraining of as few as 1\% of samples monthly. It is observed that both \drebin and \dl achieve optimal performance when the monthly retraining rate is at its maximum of 25\%, reinforcing the notion that a larger subset of retrained data is closely associated with enhanced model efficacy. It should be noted that at a monthly retraining rate of 100\%, the AL approach effectively equates to Incremental Retraining.

\begin{figure}[tp]
    \centering
    \begin{subfigure}{0.48\textwidth}
        \includegraphics[width=\linewidth]{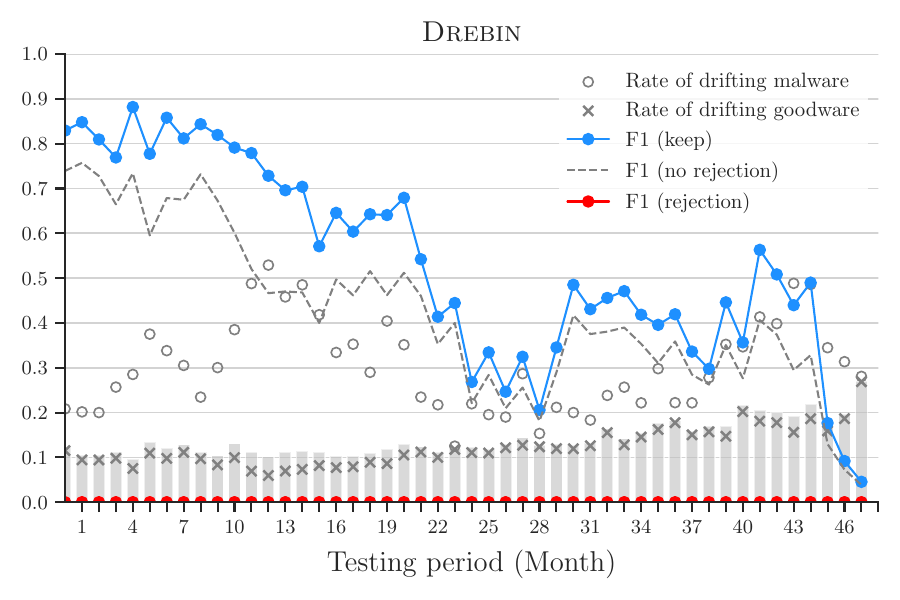}
        \label{fig:delay_transcend_drebin}
    \end{subfigure}
    \hfill
    \begin{subfigure}{0.48\textwidth}
        \includegraphics[width=\linewidth]{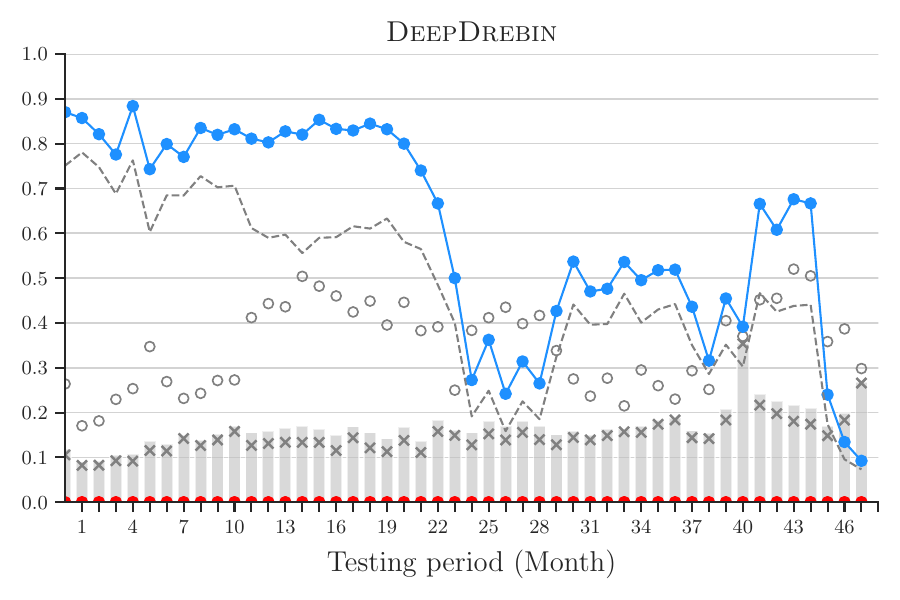}
        \label{fig:delay_transcend_dl}
    \end{subfigure}
    \vspace{-2em}
    \caption{Delay time decay: classification with \transcend \cite{barbero2022transcending} rejection.}
    \label{fig:delay_transcend}
\end{figure}

\textbf{Classification with Rejection.} As a mitigation approach, classifiers can designate decisions as "low confidence" and defer them to a later date, effectively placing uncertain predictions into a quarantine zone pending manual review~\cite{Bart:Reject}. \autoref{fig:delay_transcend} demonstrates how both \drebin{} and \dl{} perform when implementing a rejection mechanism for dubious classifications, utilizing the \transcend{} framework~\cite{Jordaney:Transcend, barbero2022transcending}. This method employs conformal evaluation theory to set thresholds that filter out the least reliable classifier decisions. Specifically, the inductive conformal evaluator (ICE) is used to pinpoint and exclude examples that deviate from expected patterns. While the original research focused on the nonconformity measure (NCM) for traditional ML algorithms, an adaptation of NCM has been developed for DNN (\dl) in this work, drawing upon the probabilistic outputs from its final SoftMax layer. Background gray histograms in the figures quantify the monthly count of quarantined objects, with cross and circle markers indicating the proportion of rejected malware and goodware, respectively.

\subsection{Analysis of Delay Methods}

To quantify performance-cost trade-offs of methods to delay time decay, we characterize the following three elements: 

{\bf Performance} (\performance{}): the performance measured in terms of AUT to capture robustness against time decay (\S~\ref{subsec:performance metrics}); 

{\bf Labeling Cost} (\labelingcost{}): the number of testing objects (if any) that must be labeled---the labeling must occur periodically (e.g., every month), and is particularly costly in the malware domain as manual inspection requires many resources (infrastructure, time, expertise, etc)---for example, Miller \etal{}~\cite{Miller:Reviewer} estimated that an average company could only manually label 80 objects per day; 

{\bf Quarantine Cost} (\quarantinecost{}): the number of objects (if any) rejected by the classifier---these must be manually verified, so there is a cost for leaving them in quarantine.

\autoref{tab:pc-tradeoff} delivers an in-depth performance-cost analysis of various delay methods applied to the \drebin and \dl malware classifiers. Using the AUT for the $F_1$-Score across two intervals - 24 and 48 months - the table outlines the implications of each method in terms of over all \labelingcost{}, \quarantinecost{} and performance gains.

The performance metrics are delineated based on two training scenarios: one with an estimated base ratio of malware in the training set ($\hat{\sigma}$) set at approximately 10\% for both classifiers, and another with the optimized ratio ($\varphi^*_{F_1}$) determined to be 55\% for \drebin and 20\% for \dl, as detailed in \autoref{tab:varphi}. The best AUT outcomes for \drebin and \dl, accentuated in the table with distinct color highlights (orange and blue, respectively) , facilitate direct comparison and underscore the impact of the tuning process.

The initial setup with no updates provides a baseline for performance, just as shown in \autoref{fig:drebin decay} in \S~\ref{subsec:reveal performance}. The training ratios with $\varphi^*_{F_1}$, however, exhibit better results for both classifiers without incurring additional costs, underscoring the importance of precise initial training/tuning settings for sustained performance.

Upon examining delay methods aimed at mitigating performance degradation over time, the tuned training ratio $\varphi^*_{F_1}$ consistently yields performance enhancements for both classifiers. Furthermore, over the 48-month period, \dl demonstrates greater resilience in maintaining performance, suggesting its robustness to temporal decay—a significant attribute for systems intended for prolonged use without continuous modification.

A closer inspection of the costs involved in \autoref{tab:pc-tradeoff} reveals that performance improvements in the $F_1$-Score are accompanied by financial considerations. The more data we label, the better retrained performance we get. For example, employing Active Learning (AL) with a 25\% relabeling rate results in a labeling cost of 50,355 samples over 48 months for both \drebin and \dl. This data prompts a careful evaluation of the balance between performance gains and the associated financial outlay. Additionally, we have to remember the costs that come from dealing with uncertain cases, which may not always be harmful but are difficult for the system to handle.

\begin{table}[t]
\centering
\small
  \caption{Performance-cost comparison of delay methods.}
        \resizebox{\columnwidth}{!}{%
  \begin{tabular}{|l||c|c||c|c||c|c||c|c||c|c||c|c|}
\hhline{~------------}
   \multicolumn{1}{c||}{} &
  \multicolumn{4}{c||}{\textbf{Costs}} &
    \multicolumn{8}{c|}{\textbf{Performance}} \\ \hhline{~------------}
 \multicolumn{1}{c||}{} &
  \multicolumn{2}{c||}{\multirow{2}{*}{\labelingcost{}(48m)}} &
    \multicolumn{2}{c||}{\multirow{2}{*}{\quarantinecost{}(48m)}} &
    \multicolumn{4}{c||}{\performance{} : AUT($F_1, 48m$)}&
    \multicolumn{4}{c|}{\performance{} : AUT($F_1, 24m$)} \\
\hhline{-~~~~--------}

  \multirow{1}{*}{\bf Delay} &
  \multicolumn{2}{c||}{} &
    \multicolumn{2}{c||}{} &
    \multicolumn{2}{c|}{$\varphi=\hat{\sigma}$} &
    \multicolumn{2}{c||}{$\varphi=\varphi^*_{F_1}$}&
    \multicolumn{2}{c|}{$\varphi=\hat{\sigma}$} &
    \multicolumn{2}{c|}{$\varphi=\varphi^*_{F_1}$} \\
\hhline{~------------}

   {\bf method} &
  \drebin{} & \dl{} &
  \drebin{} & \dl{} &
  \drebin{} & \dl{} & 
  \drebin{} & \dl{} & 
  \drebin{} & \dl{} &
  \drebin{} & \dl{}  \\ \hline \hline

No update &0&0&0&0&0.496& 0.489& \cellcolor{orange!50}0.505&\cellcolor{blue!50}  0.534&0.609&0.636&\cellcolor{orange!50}0.645&\cellcolor{blue!50} 0.667\\ \hline \hline

AL: 1\%&1,993&1,993&0&0&0.632& 0.651 & \cellcolor{orange!50} 0.655& \cellcolor{blue!50}  0.657&0.723&0.660&\cellcolor{orange!50}0.740&\cellcolor{blue!50} 0.686\\ \hline
AL: 5\%&10,052&10,052&0&0& 0.698 & 0.713 & \cellcolor{orange!50}0.722& \cellcolor{blue!50} 0.735&0.778&0.709&\cellcolor{orange!50}0.800&\cellcolor{blue!50} 0.735\\ \hline
AL: 10\%&20,127&20,127&0&0& 0.705 & 0.753 & \cellcolor{orange!50}0.738&\cellcolor{blue!50} 0.764&0.808&0.757&\cellcolor{orange!50}0.815&\cellcolor{blue!50} 0.761\\ \hline
AL: 25\%&50,355&50,355&0&0& 0.763 & 0.793 & \cellcolor{orange!50}0.789&\cellcolor{blue!50} 0.801&0.809&0.775&\cellcolor{orange!50}0.814&\cellcolor{blue!50} 0.791\\ \hline
Inc. retrain&201,490&201,490&0&0& 0.794 & 0.819 &\cellcolor{orange!50} 0.802&\cellcolor{blue!50} 0.838&0.808&0.812&\cellcolor{orange!50}0.810&\cellcolor{blue!50} 0.834\\ \hline

Rejection ($\hat{\sigma}$) &0&  0& 29,391 & 35,127 &0.537& 0.613 & -- & -- & 0.658 &  0.779 &--&--\\ \hline

Rejection ($\varphi^*_{F_1}$) &0&  0& 10,891 & 19,695 &--&--&\cellcolor{orange!50} 0.558& \cellcolor{blue!50} 0.635&--&--& \cellcolor{orange!50} 0.717 &  \cellcolor{blue!50} 0.804 \\ \hline
  \end{tabular}%
  }

  \label{tab:pc-tradeoff}

\end{table}

In the rejection scenarios using \transcend{}, the advantages of the tuning algorithm are particularly noticeable, especially for \dl. A comparison between the two rejection rows clearly shows that the application of the optimal training ratio $\varphi^*_{F_1}$ allows the classifier to more effectively quarantine dubious decisions. For \drebin, \transcend{} leads to a significant decrease in rejected samples—two-thirds less—while still improving performance: AUT($F_1$, 24m) rises from 0.658 to 0.717, and AUT($F_1$, 48m) from 0.537 to 0.558 respectively. Notably, \transcend{} even enables \dl to outperform the AUT($F_1$, 24m) of the Active Learning scenario with a 25\% budget, while quarantining fewer samples.

In conclusion, \autoref{tab:pc-tradeoff} offers not just an academic exercise but a concrete tool for the industry. It underlines the critical need for an in-depth understanding of classifier performance within actual operational contexts. This involves striking a delicate balance between analytical assessment of algorithms and the practical aspects of resource distribution for labeling and quarantining actions. The methodology outlined in this table provides a pathway to reduce costs and improve classifier performance, making it a crucial asset for the field. The challenge of creating cost-effective and sturdy classification systems is significant, yet sharing the code related to this research is expected to encourage ongoing innovation and real-world usage.

\section{Related Work}
\label{sec:related}

\textbf{Experimental Bias.} The \emph{base rate fallacy}~\cite{Axelsson:BaseRate} is a well-known experimental bias in security, which highlights the limitations of TPR and FPR as performance metrics in highly imbalanced datasets. For instance, in network intrusion detection, where most traffic is benign, even a low FPR of 1\% may correspond to millions of false positives and only thousands of true positives. In contrast, our work uncovers experimental settings that are misleading regardless of the metrics used, and that remain incorrect even if the correct metrics are employed (\S~\ref{subsec:reveal performance}). While previous works such as Sommer and Paxson~\cite{Sommer:Outside}, Rossow \etal~\cite{rossow2012prudent}, and Kouwe \etal~\cite{Van:BenchmarkingCrimes} provide valuable insights and guidelines for conducting security experiments, they do not address the specific issue of temporal and spatial bias in the Android platform. Furthermore, their guidelines would not eliminate all sources of such biases as identified in our work. For example, Rossow \etal~\cite{rossow2012prudent} evaluate the percentage of objects in previously adopted datasets that are incorrectly labeled, without assessing the impact of such errors on classifier performance. Zhou \etal~ \cite{Zhou:HPC} show that Hardware Performance Counters (HPCs) are not very effective for malware classification, but their focus is narrow, and they rely on 10-fold cross-validation in the evaluation. Finally, a recent paper published by Arp \etal~\cite{Arp:DoDont} identifies ten pitfalls for machine learning in computer security, which relates to the spatio-temporal biases we discuss. The pitfalls of sampling bias and base rate fallacy overlap with spatial bias. Along with data snooping and spurious correlations relating to temporal bias. It is important to note that these connections are in part due to the paper building off our original release of \tesseract~\cite{pendlebury2019tesseract}. However, Arp \etal~\cite{Arp:DoDont} provide the prevalence, implications, and recommendations for these pitfalls opposed to the concrete solutions provided in this work for spatio-temporal biases. 

\textbf{Time-aware Detection.} Allix \etal~\cite{Allix:Timeline} made an important contribution by evaluating malware classifiers with respect to time and demonstrating how future knowledge can inflate performance. However, they did not propose a solution for comparable evaluations and only identified constraint C1. Additionally, Allix \etal~\cite{Allix2016} investigated the differences between in-the-lab and in-the-wild scenarios and found that a greater presence of goodware leads to lower performance. In contrast, our work systematically analyzes and explains these issues, and addresses them by formalizing a set of constraints that jointly consider the impact of temporal and spatial bias. We introduce AUT as a unified performance metric for fair time-aware comparisons of different solutions, and offer a tuning algorithm to leverage the effects of training data distribution. While Miller \etal~\cite{Miller:Reviewer} identified \emph{temporal sample consistency} (equivalent to our constraint C1), they did not identify C2 or C3, which are fundamental (\S~\ref{subsec:reveal performance}). Moreover, they considered the test period to be a uniform time slot, whereas we take time decay into account. Roy \etal~\cite{Roy:ExpML} questioned the use of recent or older malware as training objects and the performance degradation in testing real-world object ratios. However, their experiments were designed without considering time, which reduces the reliability of their conclusions. Past work has highlighted some sources of experimental bias~\cite{Miller:Reviewer,Roy:ExpML,Allix:Timeline,Allix2016}, but it has given little consideration to classifiers' goals. Different scenarios may have different objectives, which may not necessarily involve maximizing $F_1$. In our work, we demonstrate the effects of different training settings on performance goals and propose an algorithm to properly tune a classifier accordingly (\S~\ref{subsec:tuning alg}). Following the publication of \cite{tesseract:poster,pendlebury2019tesseract}, numerous studies have adopted time-aware evaluations to demonstrate their performance across various temporal stages. Notable examples include \cite{xu2019droidevolver,zhang2020enhancing,barbero2022transcending,chen2023continuous}.

\textbf{Unbiased Methods from Other Domains.} The pursuit of unbiased methods is a crucial aspect in various other fields~\cite{Weiss:ReBalance,He:Imbalanced,chawla2004special}. However, since these studies do not originate from the security domain, they only focus on some aspects of spatial bias and do not consider temporal bias. Concept drift is especially problematic in Android malware compared to other applications, such as image and text classification~\cite{Jordaney:Transcend, barbero2022transcending}. Fawcett~\cite{Fawcett:Spam} highlights challenges in spam detection, where one of the challenges is similar to spatial bias, but without providing any solution. In contrast, we propose C3 to address spatial bias and demonstrate how its violation inflates performance (\S~\ref{subsec:reveal performance}). Torralba and Efros~\cite{Torralba:Unbiased} discuss dataset bias in computer vision, which is different from our security setting where there are fewer benchmarks, and the negative class (e.g., "not cat") in images can grow arbitrarily, which is less likely in the malware context. Moreno-Torres \etal~\cite{Moreno:Unifying}  systematize different types of drift and mention sample-selection bias, which resembles spatial bias, but they do not propose any solution or experiments to evaluate its impact on ML performance. Other related work emphasizes the importance of choosing appropriate performance metrics to avoid an incorrect interpretation of the results (e.g., ROC curves are misleading in an imbalanced dataset~\cite{hand2009measuring,davis2006relationship}). In this paper, we consider imbalance, propose actionable constraints and metrics, and provide tool support to evaluate the performance decay of classifiers over time.

\textbf{Summary.} Our research is motivated by several studies on bias, but none of them comprehensively address the problem in the context of evolving data where the i.i.d. assumption does not hold. While Miller \etal~\cite{Miller:Reviewer} introduced Constraint C1, it alone is insufficient to eliminate bias. It is demonstrated by other evaluations such as \cite{Mariconti:MaMaDroid}, which only enforces C1. In \S~\ref{subsec:reveal performance} of our paper, we clarify why our novel constraints C2 and C3 are fundamental and show how our AUT metric can effectively reveal the true performance of algorithms. Our approach provides counter-intuitive results that highlight the importance of considering both spatial and temporal bias when evaluating classifiers on evolving data.
\section{Discussion}
\label{sec:discussion}

We now discuss guidelines, our assumptions, and how we address limitations of our work.

{\bf Further Considerations on Temporal Consistency (Constraint C1).} One might be tempted to test how their classifier performs with respect to older data. This essentially requires to modify C1 to account for two distinct testing time frames that do not overlap with the training data one. For instance, antivirus companies might intentionally use newer threats to make sure their models can still detect older yet relevant security risks. In this case C1 can be modified to:
\begin{equation}
	\mathit{time}(s_j) < \mathit{time}(s_i) < \mathit{time}(s_k), \forall s_i \in \mathit{Tr}, \forall s_j, s_k \in \mathit{Ts}
	\label{eq:c1-modified}
\end{equation}

This raises important philosophical questions about classifier evaluation methods, representing an open research question that we suggest to address as future work.

{\bf Generalization to other security domains.}
Our \tesseract{} methodology, while demonstrated in three security domains, is broadly applicable across various ML-driven security areas for bias-free evaluations. It adapts to different domains through certain specific parameters like time granularity and test duration. This need for domain-specific settings is an inherent aspect, not a limitation, of our approach. Generally, spatio-temporal bias is likely to impact other security domains experiencing concept drift, but further research is needed for definitive conclusions. The effectiveness of \tesseract{} in understanding this bias hinges on having access to extensive timestamped datasets, realistic class ratio knowledge, and either the code or detailed information to reproduce existing baselines.

{\bf Domain-specific in-the-wild malware percentage~$\hat{\sigma}$.}
Correctly estimating the malware percentage in dataset is a challenging task and we encourage further measurement studies~\cite{lindorfer2014andradar,Perdisci:Measuring} and data sharing to obtain realistic experimental settings in different security domains.

{\bf Correct observation labels.} We assume labels in datasets we use are correct. Miller \etal~\cite{Miller:Reviewer} found that AVs sometimes change their outcome over time: some goodware may eventually be tagged as malware.  However, they also found that VirusTotal detections stabilize after one year; since we are using observations up to Dec 2018, we consider labels we used as reliable. In the future, we may integrate approaches for \emph{noisy oracles}~\cite{Du:NoisyAL}, which select only trustful observations.

{\bf Timestamps in the dataset.} It is important to consider that some timestamps in a public dataset could be incorrect or invalid. In this paper, we rely on the public AndroZoo dataset maintained at the University of Luxembourg, and we rely on the {\tt dex\_date} attribute as the approximation of an observation timestamp, as recommended by the dataset creators~\cite{Allix:AndroZoo}. We further verified the reliability of the {\tt dex\_date} attribute by re-downloading VirusTotal~\cite{VT:URL} reports for 25K apps\footnote{We downloaded only 25K VirusTotal reports (corresponding to about 10\% of our dataset) due to restrictions on our VirusTotal API usage quota.} and verifying that the {\tt first\_seen} attribute always matched the {\tt dex\_date} within our time span.
In general, we recommend performing some sanitization of a timestamped dataset before performing any analysis on it: if multiple timestamps are available for each object, consider the most reliable timestamp you have access to (e.g., the timestamp recommended by the dataset creators, or the VirusTotal's {\tt first\_seen} attribute) and discard objects with ``impossible'' timestamps (e.g., with dates which are either too old or in the future), which may be caused by incorrect parsing or invalid values of some timestamps. To improve trustworthiness of the timestamps, one could verify whether a given object contains time inconsistencies or features not yet available when the app was released~\cite{Li:MoonlightBox}. We encourage the community to promptly notify dataset maintainers of any date inconsistencies.
 In the \tesseract{}'s project website (\autoref{sec:availability}), we will maintain an updated list of timestamped datasets publicly available for the security community.

{\bf Resilience of malware classifiers.} In our study, we analyze both traditional ML approach \drebin and DNN approach \dl{}. One could argue that other classifiers may show consistently high performance even with space-time bias eliminated. And this should indeed be the goal of research on malware classification. \tesseract{} provides a mechanism for an unbiased evaluation that we hope will support this kind of work.

{\bf Adversarial ML.} Adversarial ML focuses on perturbing training or testing observations to compel a classifier to make incorrect predictions~\cite{Biggio:Wild}. Both relate to concepts of \textit{robustness} and one can characterize adversarial ML as an artificially induced worst-case concept drift scenario. While the adversarial setting remains an open problem, the experimental bias we describe in this work---endemic in Android malware classification---must be addressed prior to realistic evaluations of adversarial mitigations.

\section{Availability}
\label{sec:availability}

We make \tesseract{}'s code and data available to the research community to promote the adoption of a sound and unbiased evaluation of classifiers. The \tesseract{} project website with instructions to request access is at \mbox{\sf https://github.com/s2labres/tesseract-ml/}, and all experiments related code can be found at \mbox{\sf https://github.com/s2labres/tesseract-journal-experiments/}. We will also keep maintaining an updated list of publicly available security-related datasets with timestamped objects.

\section{Conclusions}
\label{sec:conclusions}

 We have identified novel temporal and spatial biases in ML-driven malware detection, and proposed novel constraints, metrics and tuning algorithm to address such issues. We have built and released \tesseract{} as an open-source tool that integrates our methods. We have shown the two experimental biases commonly exist in detection tasks, and how \tesseract{} can reveal the real performance of malware classifiers that remain hidden in wrong experimental settings in a non-stationary context. \tesseract{} is fundamental for the correct evaluation and comparison of different solutions, in particular when considering mitigation strategies for time decay.

 \tesseract{} offers valuable insights for both researchers and practitioners. It enables evaluation of a classifier's inherent robustness to performance decay using a baseline AUT performance metric, highlighting true performance and sometimes surprising results. This is particularly useful in deployment scenarios where frequent retraining is not feasible due to financial or computational constraints. Our experiments reveal that even with retraining, classifiers may not remain consistent over time, and \tesseract{} assists in eliminating spatio-temporal bias in evaluations, allowing researchers to focus on developing more robust algorithms. Our experiments also indicate the necessity of regular retraining and tuning for sustained performance. Additionally, \tesseract{} supports various time granularities for analysis and allows different observation time windows for evaluation, enabling researchers to assess and compare the robustness of algorithms over specific time frames, such as the upcoming three months, ensuring bias-free comparisons and consideration of time decay.

We envision that future work on malware classification will use \tesseract{} to produce realistic, comparable and unbiased results. We encourage the community to adopt \tesseract{} to deeply evaluate the impact of temporal and spatial bias in any security domain where concept drift appears.

\newpage 
\bibliographystyle{plain}
\bibliography{ref}

\appendix
\section{Appendix}
\subsection{Algorithm Hyperparameters}
\label{app:hyperparameters}

In the study of malware detection algorithms, hyperparameters play a critical role in defining the behavior and performance of the classification models. This section details the specific hyperparameters utilized for the two algorithms tested: \drebin and \dl.

The \drebin algorithm is instantiated using a linear Support Vector Machine (SVM) from the \sklearn library, specifically employing {\it sklearn.svm.LinearSVC}. In alignment with the original \drebin study, the SVM was fine-tuned with a regularization parameter C set to 1. This choice reflects a balance between the correct classification of training examples and the maximization of the decision function’s margin. The dual parameter's automatic setting ensures the algorithm optimally chooses between the primal and dual problem formulations, contingent on the dataset's dimensionality. The solver's iterations were limited to 5000 to establish a convergence threshold, reflecting a practical tolerance for the training data's optimization process.

This hyperparameter configuration is a deliberate effort to replicate the original \drebin paper's findings. Despite slight modifications to accommodate the current study's dataset, this setup achieved a commendable 10-fold cross-validated $F_1$ score of around 0.91, a figure modestly trailing the original performance, thereby validating the reproducibility and robustness of the approach.

The \dl{} model represents a deep learning approach, architected with a multi-layer sequential design. It comprises an input layer, succeeded by two hidden layers with 200 neurons each. Post each hidden layer, a ReLU (Rectified Linear Unit) function is applied, injecting non-linearity, coupled with a dropout strategy at a rate of 0.5 to mitigate overfitting. The terminal layer of the network is binary, outputting two values for the classification task.

Training hyperparameters for \dl{} were meticulously set: the model underwent 10 epochs, ensuring thorough exposure to the training dataset. The batch size was chosen as 64, balancing computational load with frequency of parameter updates. A learning rate of 0.05 was selected to navigate the parameter space effectively. Stochastic Gradient Descent (SGD) serves as the optimizer, steering the loss function towards an optimal minimum.

In adherence to the protocols in~\cite{Papernot:ESORICS}, \dl{} was re-implemented using \torch{}. Initial inputs were aligned with those utilized by \drebin{}. We mirrored the neural network architecture that demonstrated optimal results as per~\cite{Papernot:ESORICS}. Yet, we encountered a lack of comprehensive detail on hyperparameter optimization in the source material. Our refined implementation, however, culminated in a marginally elevated $F_1$ score within a 10-fold cross-validation framework (\S~\ref{subsec:reveal performance}). We postulate that this increment stems from the original research's optimization for the Accuracy metric~\cite{Bishop:ML}, which can be deceptive in imbalanced datasets~\cite{Axelsson:BaseRate}—a common characteristic in the Android domain, dominated by goodware.

\subsection{Symbol Table}
\label{app:symbol}

\autoref{tab:symbol} is a legend of the main symbols used throughout this paper to improve readability.

\begin{table}[h!]
\caption{Symbol table.}
\centering
\small
{
\footnotesize
  \begin{tabular}{>{\centering\arraybackslash}m{0.2\textwidth}|m{0.6\textwidth}}
\centering
  \textbf{Symbol} & \textbf{Description} \\
  \midrule
  gw & Short version of goodware. \\  \hline
  mw & Short version of malware. \\  \hline
  ML & Short version of Machine Learning. \\  \hline
  \dataset{} & Labeled dataset with malware (mw) and goodware (gw). \\  \hline
  $\mathit{Tr}$ & Training dataset. \\  \hline
  $W$ & Size of the time window of the training set (e.g., 1 year). \\  \hline
  $\mathit{Ts}$ & Testing dataset. \\  \hline
  $S$ & Size of the time window of the testing set (e.g., 2 years). \\  \hline

  $TP, FP, TN, FN$ & Stand for True Positive, False Positive, True Negative, \\ &and False Negative, respectively \\  \hline
  
  $\Delta$ & Size of the test time-slots for time-aware evaluations (e.g., months). \\  \hline
  $\mathbb{P}$ & Performance target of the tuning algorithm in \S~\ref{subsec:tuning alg}; it can be $F_1$-Score ($F_1$), Precision ($Pr$) or Recall ($Rec$). \\ \hline
  AUT($\mathbb{P}$,$N$) & Area Under Time, a new metric we define to measure performance over time decay and compare different solutions (\S~\ref{subsec:performance metrics}). It is always computed with respect to a performance function $\mathbb{P}$ (e.g., $F_1$-Score) and $N$ is the number of time units considered (e.g., 24 months)\\ \hline
  $\tau$ & Size of AUT observation time window. \\  \hline
  $\hat{\sigma}$ & Estimated percentage of malware (mw) in the wild. \\ \hline
  $\varphi$ & Percentage of malware (mw) in the training set. \\  \hline
  $\delta$ & Percentage of malware (mw) in the testing set. \\  \hline
  
  $\varphi^*_\mathbb{P}$ & Percentage of malware (mw) in the training set, to improve performance $\mathbb{P}$ on the malware (mw) class (\S~\ref{subsec:tuning alg}). \\  \hline
\errorrate{} & Error rate (\S~\ref{subsec:tuning alg}). \\  \hline
\errorratemax{} & Maximum error rate when searching $\varphi^*_\mathbb{P}$ (\S~\ref{subsec:tuning alg}). \\  \hline
 $\Theta$ & Model learned after training a classifier. \\  \hline
 \labelingcost{} & Labeling cost. \\  \hline
 \quarantinecost{} & Quarantine cost. \\  \hline
 \performance{} & Actual performance; depending on the context, it can refer to AUT with $F_1$ or $Pr$ or $Rec$. \\  \hline
  \bottomrule
  \end{tabular}
}

\label{tab:symbol}

\end{table}

\subsection{Cumulative Plots for Time Decay}
\label{app:cml}
\begin{figure}[t]
\centering
    \includegraphics[width=0.85\textwidth]{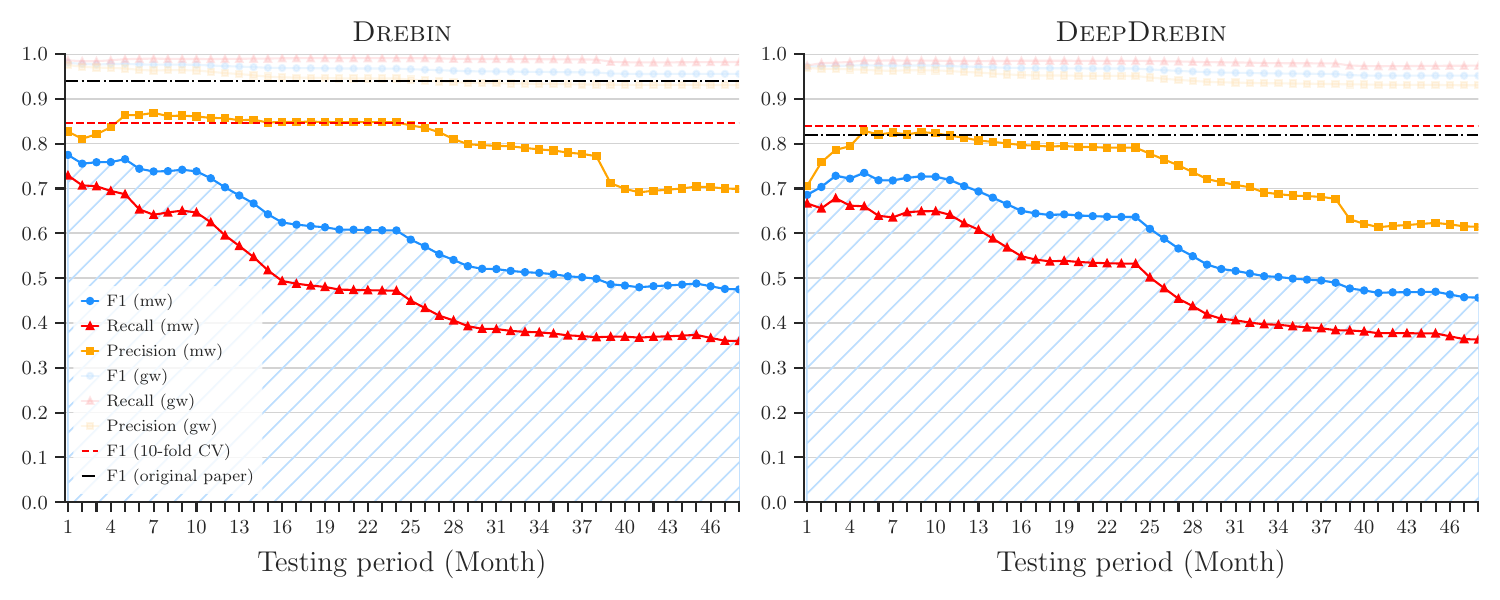}
	
	\vspace{-1em}
	\caption{Performance time decay with cumulative estimate for \drebin{} and \dl{}. Testing distribution has $\delta=10\%$ malware, and training distribution has $\varphi=10\%$ malware.}
	\label{fig:decay-cum}
 
\end{figure}

\autoref{fig:decay-cum} shows the cumulative performance plot defined in~\S~\ref{subsec:performance metrics}. This is the cumulative version of~\autoref{fig:drebin decay} on the Android dataset.

\subsection{\tesseract Implementation}
\label{sec:artifact}

Our Python library, \tesseract, incorporates our constraints, metrics, and algorithms, and is designed for seamless integration with standard workflows. Its API, closely aligned with and fully compatible with \sklearn~\cite{scikit-learn} and \torch~\cite{pytorch}, ensures familiarity for users of these libraries. A detailed overview of \tesseract's core modules is available in~\cite{tesseract:poster}.

\textbf{loader.py} is designed to transform the input dataset into a structure compatible with \tesseract. It outputs a vectorized matrix \lstinline{X}, a label array \lstinline{y}, and an associated \lstinline{datetime} array \lstinline{t}. This transformation can be executed by calling the \lstinline{load_features()} function. Additionally, the file contains the feature selection function (\lstinline{feature_reduce()}), utilized in \S~\ref{subsec:select}, for efficient feature processing.

\textbf{temporal.py} enhances traditional machine learning frameworks that utilize input features \lstinline{X} and target variables \lstinline{y} by incorporating \lstinline{datetime} objects \lstinline{t}. This addition enables time-sensitive data operations, such as dataset partitioning which adheres to temporal constraints C1 and C2, with functions like \lstinline{time_aware_partition()} and \lstinline{time_aware_train_test_split()}. A \lstinline{granularity} parameter is also provided, allowing the specification of the evaluation time window.

{\textbf{spatial.py}} This module allows the user to alter the proportion of the positive class in a given dataset. \lstinline{downsample_set()} can be used to simulate the natural class distribution $\hat{\sigma}$ expected during deployment or to tune the performance of the model by over-representing a class during training. To this end we provide an implementation of Algorithm~\ref{alg:varphi} (\S~\ref{subsec:tuning alg}) for finding the optimal training proportion $\varphi^*$ (\lstinline{search_optimal_train_ratio()}). This module can also assert that constraint C3 (\S~\ref{subsec:constrains}) has not been violated.

{\textbf{metrics.py}} As \tesseract aims to encourage comparable and reproducible evaluations, we include functions for visualizing classifier assessments and deriving metrics such as the accuracy or total errors from slices of a time-aware evaluation. Importantly we also include \lstinline{aut()}, with optional parameter granularity ($\delta$) and observation time window ($\tau$), for computing the AUT (\S~\ref{subsec:performance metrics}) for a given metric ($F_1$, Precision, Recall, etc.) over a given time~period.

{\textbf{evaluation.py}} Here we include the \lstinline{predict()} and \lstinline{fit_predict_update()} functions that accept a classifier, dataset and set of parameters (as defined in \S~\ref{subsec:constrains}) and return the results of a time-aware evaluation (\S~\ref{subsec:reveal performance}) performed across the chosen periods.

{\textbf{selection.py \textnormal{\textit{and}} rejection.py}} For extending the evaluation to testing model update strategies, these modules provide hooks for novel query and reject strategies to be easily plugged into the evaluation cycle. We already implement many of the methods discussed in \S~\ref{sec:delay} and include them with our release. We hope this modular approach lowers the bar for future researches.

\end{document}